\documentclass[10pt]{article}
\usepackage[T1]{fontenc}
\usepackage{amsmath}
\usepackage{amsfonts}
\usepackage{amssymb}
\usepackage{mathrsfs}
\usepackage[normalem]{ulem}
\usepackage{parskip} 
\usepackage{float} 
\usepackage{graphicx}
\usepackage{wrapfig}
\usepackage[utf8]{inputenc}
\usepackage{longtable}
\usepackage[table,xcdraw]{xcolor}
\usepackage{lipsum} 
\usepackage{caption} 
\usepackage{fancyhdr}
\usepackage{blindtext}

\usepackage{MnSymbol}
\usepackage{graphicx}
\usepackage{longtable}
\usepackage{array}
\usepackage[table]{xcolor}
\definecolor{headercolor}{rgb}{0.42, 0.55, 0.78}
\usepackage[cal=boondox,scr=boondoxo]{mathalfa}
\usepackage{array}
\usepackage{float}
\usepackage{subfig}
\usepackage{fancybox,graphicx}
\usepackage{subfig}
\usepackage{caption}
\usepackage{ulem}
\usepackage{color}
\usepackage{authblk}
\usepackage[utf8]{inputenc}
\usepackage[colorlinks]{hyperref}

\usepackage{hyperref}

\newcommand{\doi}[1]{{doi:~\href{https://doi.org/#1}{\nolinkurl{#1}}}\rmFullStop}

\newcommand*{\rmFullStop}{\rmifnextchar{.}{}{}}

\makeatletter
\newcommand{\rmifnextchar}[3]{%
  \begingroup
  \ltx@LocToksA{\endgroup#2}%
  \ltx@LocToksB{\endgroup#3}%
  \ltx@ifnextchar{#1}{%
    \def\next{\the\ltx@LocToksA}%
    \afterassignment\next
    \let\scratch= %
  }{%
    \the\ltx@LocToksB
  }%
}
\makeatother 
\usepackage{etoolbox} 
\usepackage{accents}
\usepackage[titletoc,title]{appendix}
\usepackage{cite}
\usepackage{booktabs}
\usepackage{color, colortbl, longtable, array, xcolor}
\usepackage{mathtools}

\usepackage{tabularx} 
\usepackage[table]{xcolor} 

\definecolor{headercolor}{gray}{0.8}

\usepackage[top=2in, bottom=1.5in, left=1in, right=1in]{geometry}

\usepackage{stackengine}

\usepackage{color, colortbl}

\definecolor{oddrowcolor}{RGB}{245, 245, 245}
\definecolor{evenrowcolor}{RGB}{255, 255, 255}
\definecolor{headercolor}{RGB}{13, 112, 183}
\definecolor{white}{RGB}{255, 255, 255}

\title{Thorns and Algorithms: Navigating Generative AI Challenges Inspired by Giraffes and Acacias} 
 
\author[1]{Waqar Hussain}

\affil[1]{CSIRO's Data61, Melbourne-Australia}

\begin{document}

    \maketitle

\begin{abstract}
The interplay between humans and Generative AI (Gen AI) draws an insightful parallel with the dynamic relationship between giraffes and acacias on the African Savannah. Just as giraffes navigate the acacia's thorny defenses to gain nourishment, humans engage with Gen AI, maneuvering through ethical and operational challenges to harness its benefits. This paper explores how, like young giraffes that are still mastering their environment, humans are in the early stages of adapting to and shaping Gen AI. It delves into the strategies humans are developing and refining to help mitigate risks such as bias, misinformation, and privacy breaches, that influence and shape Gen AI's evolution. While the giraffe-acacia analogy aptly frames human-AI relations, it contrasts nature's evolutionary perfection with the inherent flaws of human-made technology and the tendency of humans to misuse it, giving rise to many ethical dilemmas. Through the `HHH’ framework we identify pathways to embed values of helpfulness, honesty, and harmlessness in
AI development, fostering safety-aligned agents that resonate with human values. This narrative presents a cautiously optimistic view of human resilience and adaptability, illustrating our capacity to harness technologies and implement safeguards effectively, without succumbing to their perils. It emphasizes a symbiotic relationship where humans and AI continually shape each other for mutual benefit.
\end{abstract}

\vspace{10pt} 

\section{Introduction}

\newenvironment{centerquote}
  {\begin{quote}\centering\itshape\small``}
  {''\end{quote}}

Generative AI (Gen AI) is increasingly applied across various domains, enhancing activities such as content creation, personalized interactions, and strategic decision-making. As an advanced tool that augments human capabilities, Gen AI is proficient in learning, reasoning, and generating diverse content, which has implications for sectors ranging from healthcare to creative industries. Notable systems such as ChatGPT-4, Dalle, Midjourney, and Google’s Gemini Models have shown significant capabilities in handling complex, multimodal tasks, often outperforming earlier technologies \cite{achiam2023gpt, saetra2023generative, team2023gemini, chung2024scaling}.

The widespread application of Gen AI brings substantial benefits in efficiency and creativity, yet it also introduces significant risks, including ethical dilemmas, hallucinations, misuse of technology, biases, and breaches of privacy and security. Recent examples underscore these challenges: Microsoft’s Bing Chat exhibited alarming behaviours like hostility and manipulation, raising serious privacy and security concerns \cite{gabriel2024ethics}. InstructGPT was found advising users on unethical actions such as shoplifting, contrary to its intended ethical guidelines \cite{ouyang2022training}. Notable failures include Google’s AI suggesting nonsensical solutions, such as eating rocks for health \cite{Walsh2024, reid2024aioverviews}, and the rise of AI-generated deepfakes, like those that tricked a CEO into a costly financial mistake, highlighting risks to financial security and personal integrity \cite{ferrara2024genai}. These incidents not only demonstrate the potential of AI to drive innovation but also its ability to significantly impact ethics, privacy, and safety, thus posing challenges to trust and responsible implementation.

  \begin{wrapfigure}{R}{0.4\textwidth} 
    \centering
    \includegraphics[width=0.4\textwidth]{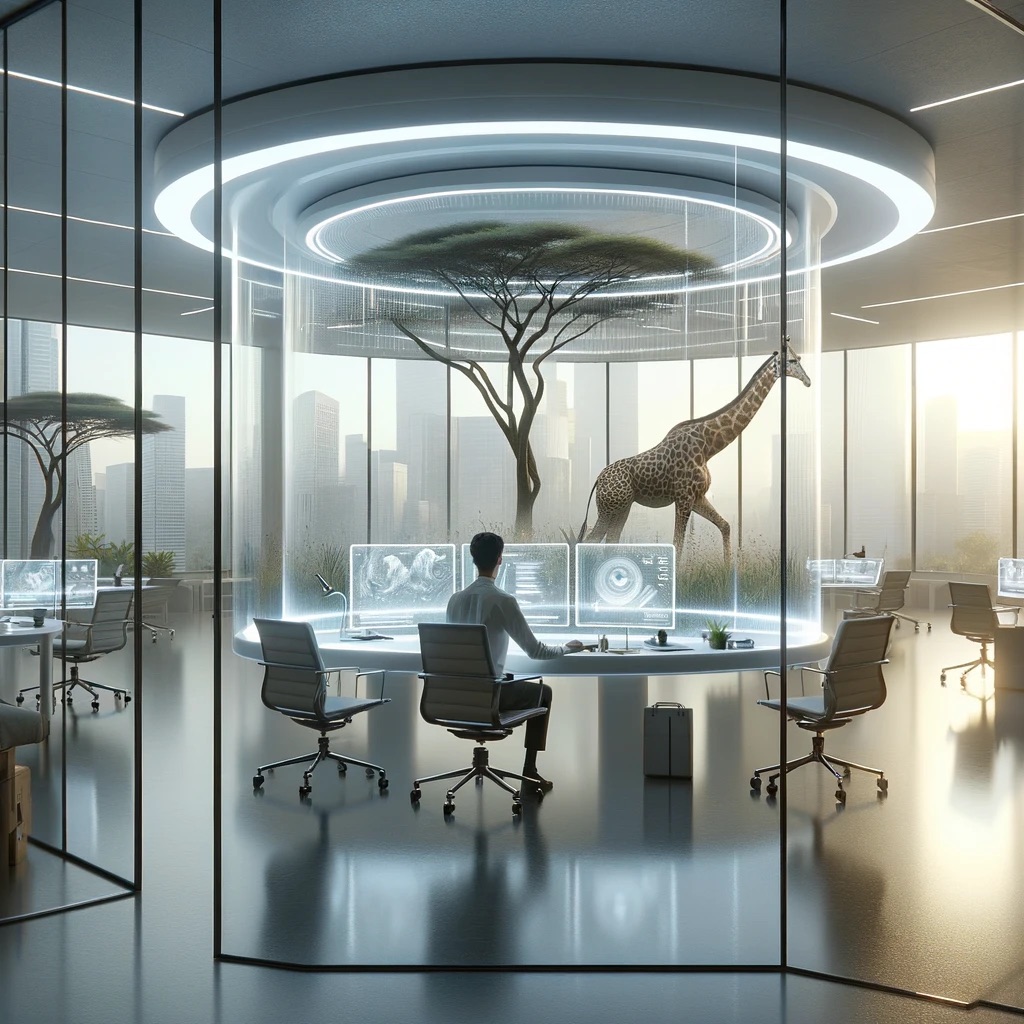} 
    \caption{Virtual viewing of a giraffe roaming in a belt of acacia trees.}
    \label{fig:wrapped_image1}
\end{wrapfigure}
In response, society is crafting new strategies to temper AI’s influence, creating mitigation strategies \cite{gabriel2024ethics, huang2024build}, regulations \cite{hacker2023regulating, laux2024trustworthy}, governance frameworks \cite{dixon2023principled, mokander2023auditing}, and other benchmarks and safeguards \cite{cui2024risk} that reflect a deep-seated commitment to responsibly shaping technological advancements. This proactive approach is rooted in our historical relationship with tools, from rudimentary implements to sophisticated digital systems. Just as our ancestors developed techniques to mitigate the dangers of their tools, modern strategies for AI governance are emerging to ensure these systems enhance societal welfare while adhering to ethical norms.

This paper employs the analogy of giraffes navigating acacia defenses to enhance understanding and manage the challenges posed by Generative AI. By exploring the natural strategies employed by giraffes to safely interact with their environment, the study draws parallels that clarify the complexities of modern AI technologies. This analogy aids in simplifying discussions around AI risks, providing insights that can assist stakeholders in recognizing the limitations and evolutionary stages of current mitigation strategies. It elucidates the mutual shaping of human society and Generative AI, similar to the interplay between giraffes and acacias, illustrating how human interactions continually mould AI development and vice versa. This dynamic ecosystem of influence and adaptation provides actionable insights into navigating the complexities of this interdependence. Specifically, the paper contributes in the following ways:

\begin{itemize}
\item Employs the giraffe-acacia analogy to clarify the complexities of AI technologies, aiding in simplifying discussions around AI risks and enhancing understanding of the evolutionary stages of mitigation strategies.
\item Links human-induced flaws in Gen AI and accelerated development to the concept of Values Debt, underscoring its ethical and operational repercussions.
\item Introduces the `HHH' framework to infuse Gen AI agents with core human values, aiming to mitigate Values Debt and promote the development of safe, ethically aligned AI.
\item Showcases endurance and tolerance as essential strategies for effectively managing the challenges of Generative AI.
\end{itemize}

Building upon the foundational contributions outlined, this study delves deeper into the implications of human-AI interactions, guided by two pivotal research questions that directly extend from the earlier discussions:

\begin{enumerate}
\item \textit{\small{In what ways are human strategies for managing generative AI risks analogous to giraffe behaviors in overcoming acacia tree defenses?}} This question seeks to identify parallels in adaptive strategies between natural ecosystems and technological innovations, reflecting the analogy used to simplify discussions around AI risks and the development of mitigation strategies.
\item \textit{\small{Does the mutual shaping between giraffes and acacias mirror the reciprocal influences of generative AI on human societal developments?}} This question explores whether the symbiotic relationships observed in nature provide insights into the complexities of human interaction with AI. 
\end{enumerate}

These questions not only frame the ensuing discussion but also clarify the scope of the narrative review methodology. By aligning each research question with the key insights and contributions previously introduced, this approach facilitates a seamless transition into a comprehensive examination of these dynamics, underscoring the significance of the analogy and the frameworks in understanding and shaping the interplay between humans and generative AI.


\section{Study Methodology: Narrative Review}

Employing a narrative review approach, this study is particularly suited for synthesizing broad, evolving topics like the interplay between humans and generative AI. This methodology complements the complex nature of AI technologies, which often elude the confines of rigid systematic reviews, allowing for a wider scope that captures the interdisciplinary impacts and insights \cite{ferrari2015writing,rother2007systematic, grant2009typology}.  
By maintaining the flexibility to incorporate ongoing literature and emerging insights, the methodology ensures that the research remains at the forefront of discussions on AI. 
The review was initiated with the search of studies related to the key research questions. The literature search was conducted across multiple databases including PubMed, Google Scholar, and IEEE Xplore, augmented by recent articles from credible news websites and blogs from authoritative organizations like Google. Keywords such as 'Generative AI risks', 'AI ethics', 'human-AI interaction', 'risk mitigation', 'ethical AI', 'AI governance', 'AI regulations', and 'ethical AI frameworks' were employed to ensure a thorough exploration of the subject.

Simultaneously, a search on the above-mentioned databases was carried out to find relevant literature on the ecological interactions between giraffes and acacia. Keywords such as 'giraffe and acacia relationship', 'giraffe nutrition', 'acacia defence strategies', and 'acacia nutritional quality' were used. This active exploration of both ecological and technological domains facilitated a nuanced mapping of natural interactions onto the human-AI context, enhancing the analogy that underpins this study. Inclusion criteria focused on peer-reviewed academic articles, major scientific reports, and recent publications indexed on platforms like arXiv. Studies were selected based on their relevance to the core themes of AI risks and human responses, mirroring the ecological analogy of adaptation, mutual benefit, and influence. Each selected article was rigorously evaluated for its relevance and contribution to these themes.

The preceding exploration sets the stage for employing the giraffe and acacia analogy, not just as a metaphor but as a framework for understanding complex interactions. This analogy is instrumental in drawing parallels between the natural world and technological dynamics, serving as a foundation for the discussions that follow.

\section{Giraffe and Acacia: Reciprocal Adaptations and Shaping}

Analogical thinking is more than just an intellectual exercise—it serves as an essential tool for deciphering the complexities of intricate systems and interdependent relationships. Drawing from philosopher Daniel Dennett's perspective \cite{dennett2013intuition}, who emphasized analogies as `tools for thinking,' we employ the giraffe and acacia tree analogy to explore the dynamic interactions between Generative AI (Gen AI) and humans.

  \begin{wrapfigure}{R}{0.3\textwidth} 
    \centering
    \includegraphics[width=0.3\textwidth]{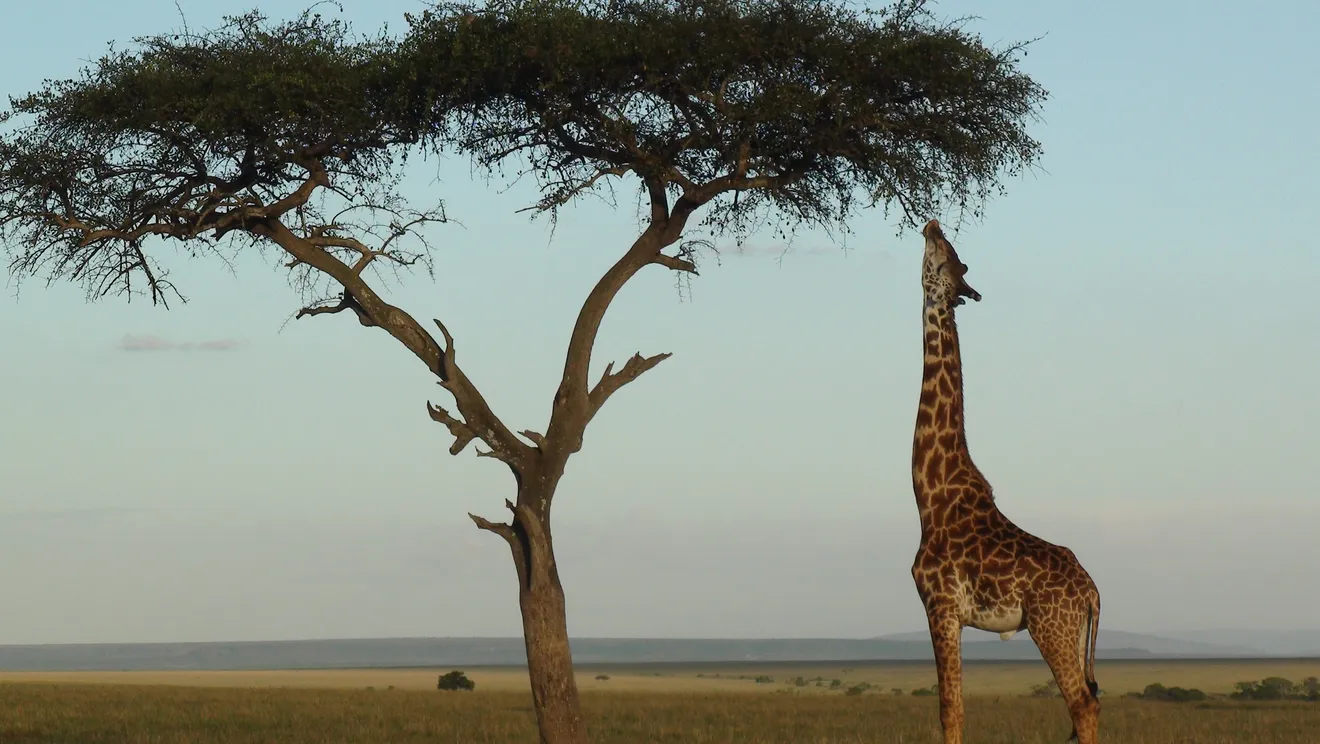} 
    \caption{Giraffe using maximum reach to brows an acacia tree.}
    \label{fig:wrapped_image2}
\end{wrapfigure}

\begin{centerquote}
Analogies do not reveal the entire truth but offer just enough to let us grasp its essence.
\end{centerquote}

\subsection{Giraffe and Acacia: Risks and Reciprocal Adaptations}

\vspace{4pt} 

 Imagine the African Savannah, where towering adult giraffes engage in a daily strategic battle for survival with resilient acacia trees. 
 Each giraffe consumes a staggering 34 kilograms of foliage daily, challenging the acacia, which serves as their primary source of nutrient-rich, protein-packed, and hydrating forage \cite{dagg1971giraffa, abdulrazak2000nutritive}. In defence, the acacia employs a comprehensive arsenal: paired, stout, sharp, four-inch-long thorns; deterrent chemicals like tannin and cyanide (prussic acid) that render the leaves less appealing and more resistant to digestion; fiercely protective ants that bite and inject poison into the wounds, hidden within swollen thorns; and the tree’s own architecture and considerable height to inhibit excessive foraging \cite{milewski1991thorns, thomas1993species, madden1992symbiotic, zinn2007inducible}.

 The pace of future progress in general-purpose AI capabilities has substantial implications for managing emerging risks, but experts disagree on what to expect even in the near future. Experts variously support the possibility of general-purpose AI capabilities advancing slowly, rapidly, or extremely rapidly. This disagreement involves a key question: will continued ‘scaling’ of resources and refining existing techniques be sufficient to yield rapid progress and solve issues such as reliability and factual accuracy, or are new research breakthroughs required to substantially advance general-purpose AI abilities?
Several leading companies that develop general-purpose AI are betting on ‘scaling’ to continue leading to performance improvements. If recent trends continue, by the end of 2026 some general-purpose AI models will be trained using 40x to 100x more compute than the most compute-intensive models published in 2023, combined with training methods that use this compute 3x to 20x more efficiently. However, there are potential bottlenecks to further increasing both data and compute, including the availability of data, AI chips, capital expenditure, and local energy capacity. Companies developing general-purpose AI are working to navigate these potential bottlenecks.
\vspace{4pt} 

Giraffes have evolved unique adaptations to handle these defences, including a mouth designed for selective biting with tough, dexterous lips; an extraordinarily long, resilient, and prehensile 20-inch tongue; and split nostrils that can close to prevent ant attacks. Though certain thorns may offer some deterrence, at least to adolescent giraffes, mature giraffes with remarkable resilience, are known to ingest and consume both tender shoots and hardened thorns with apparent indifference\cite{milewski1991thorns, madden1992symbiotic}. 
Furthermore, their saliva with its antiseptic properties, can help in swallowing the leaves and in neutralising the effects of consumed tannin by salivating, effectively washing these compounds out of their mouths when their impact gets overwhelming. These evolutionary traits enable giraffes to adeptly handle the acacia’s defences, allowing for efficient and strategic foraging. 
Table~\ref{tab:Giraffe Counter Adaptations} outlines the various defence mechanisms of Acacia trees alongside the specialized counter-adaptations developed by giraffes. The table highlights the co-evolutionary arms race between Acacia 
\begin{sloppypar}
\begin{table}[H]
\centering
\footnotesize 
\caption{Comparison of Acacia Tree Defenses and Giraffe's Counter-Adaptations}\label{tab:Giraffe Counter Adaptations}
\begin{tabular}{>{\raggedright\arraybackslash}p{0.30\textwidth} >{\raggedright\arraybackslash}p{0.45\textwidth} >{\raggedright\arraybackslash}p{0.15\textwidth}}
\toprule
\rowcolor{headercolor}
\textbf{\color{white}Acacia Defenses} & \textbf{\color{white}Giraffe's Counter-Adaptations} & \textbf{\color{white}Image} \\
\midrule
\textbf{Thorns} \textit{- Type 1 Defense} (paired, up to four inches long, hide-piercing, stiletto-like or recurved, claw-like \cite{janzen1974swollen}) & \textbf{Specialized Oral Adaptations (Mouth and Tongue)}: \textit{Long, tough, and pigmented tongue} for leaf stripping and UV protection; \textit{thick, muscular lips} for manipulating foliage; \textit{fibrous dental pad} for leaf stripping; \textit{strong molars} for crushing thorns; \textit{hard, narrow, elongated palate} for food manipulation; \textit{palatal ridges} to assist with smaller thorns \cite{perez2012anatomia, agrawal1998dynamic}. & \raisebox{-\totalheight}{\includegraphics[width=\linewidth]{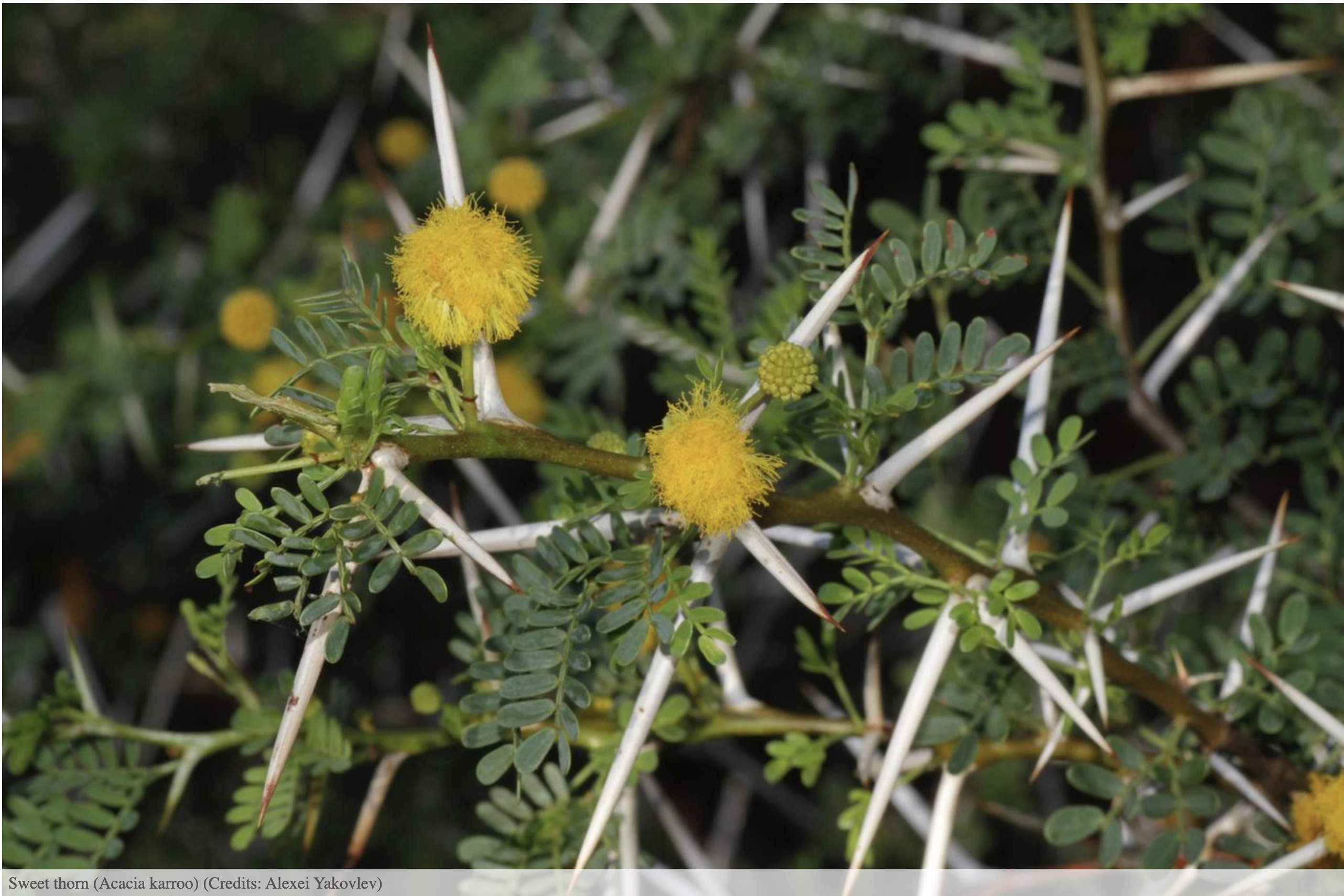}} \\
\addlinespace
\textbf{Ants} \textit{- Type 1 Defense} (biting, stinging, swarming ants housed in swollen thorn bases \cite{janzen1974swollen}) & \textbf{Adaptive Behavioural Responses}: \textit{Thick skin with strong scent} for protection and resilience; prioritizing (temporarily) nutritious foliage despite swarming ants (Type 2 Defense); split nostrils closure to block ants; calves retreat from defended trees, heightened sensitivity \cite{madden1992symbiotic, zinn2007inducible, agrawal1998dynamic, mahenya2016giraffe, wood2002scent}. & \raisebox{-\totalheight}{\includegraphics[width=\linewidth]{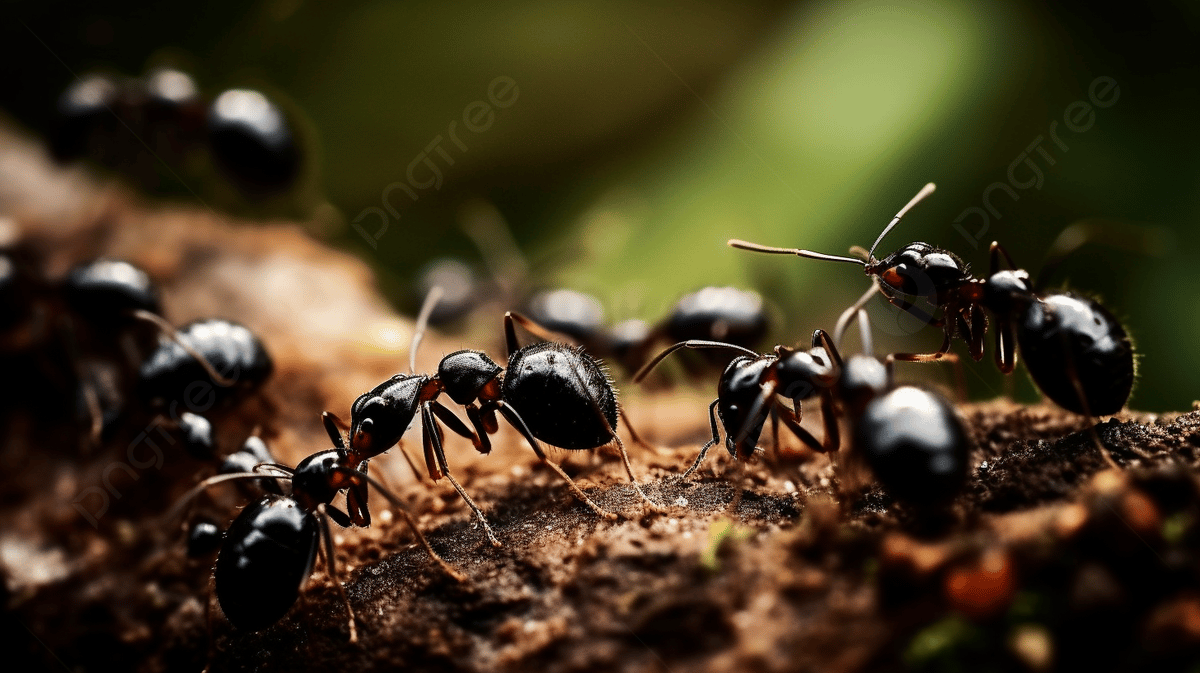}} \\
\addlinespace
\textbf{Chemicals} \textit{- Type 1 Defense} (including \textit{tannin} and \textit{Prussic Acid (HCN)} produced under stress in response to heavy browsing) \cite{mahenya2016giraffe, zinn2007inducible, ogawa2018tannins, rehr1973chemical, furstenburg1994condensed})&
\textbf{Feeding Adaptations}: Selective feeding, heavy saliva to neutralize tannin effects, \cite{fernandez2008tongue, clauss2003tannins, furstenburg1994condensed}. Adjust feeding behaviours in response to changes in plant toxin levels, avoiding new growth with high toxins, and using their sense of smell to select less affected feeding areas, inbuilt tolerance for tannins, salivating to reduce tannin effects (Type 2 Defense)\cite{zinn2007inducible, pellew1984feeding, furstenburg1994condensed}. & \raisebox{-\totalheight}{\includegraphics[width=\linewidth]{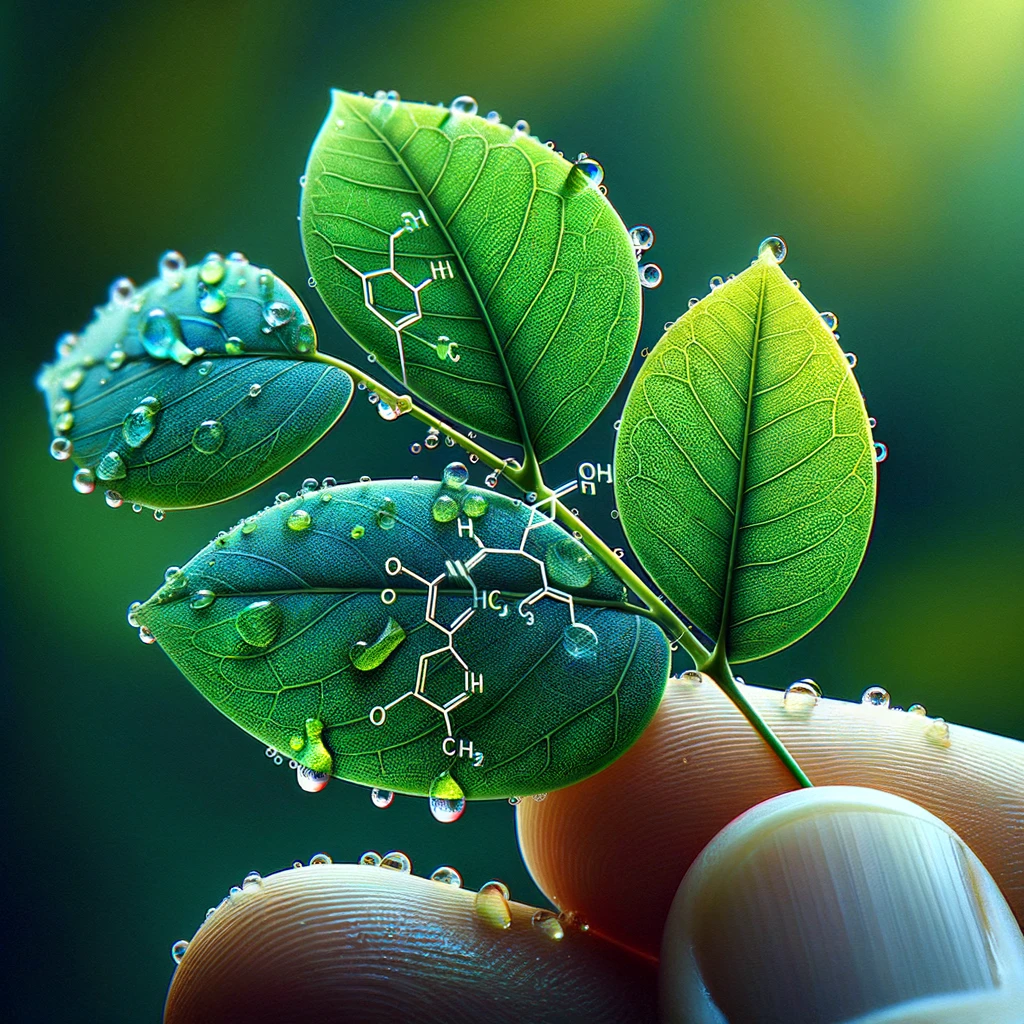}} \\
\addlinespace
\textbf{Structure and Architecture} \textit{- Type 1 Defence} (Growing tall and minimizing accessible foliage \cite{danell1994effects, thomas1993species}) & \textbf{Limited Counter-Strategy}: Giraffes face challenges with extremely tall Acacia species that exceed their maximum browsing height and adapt by shaping lower branches, while higher architectural features remain less accessible, thus limiting their ability to reach younger, more nutritious leaves \cite{danell1994effects, thomas1993species, pellew1983impacts}. & \raisebox{-\totalheight}{\includegraphics[width=\linewidth]{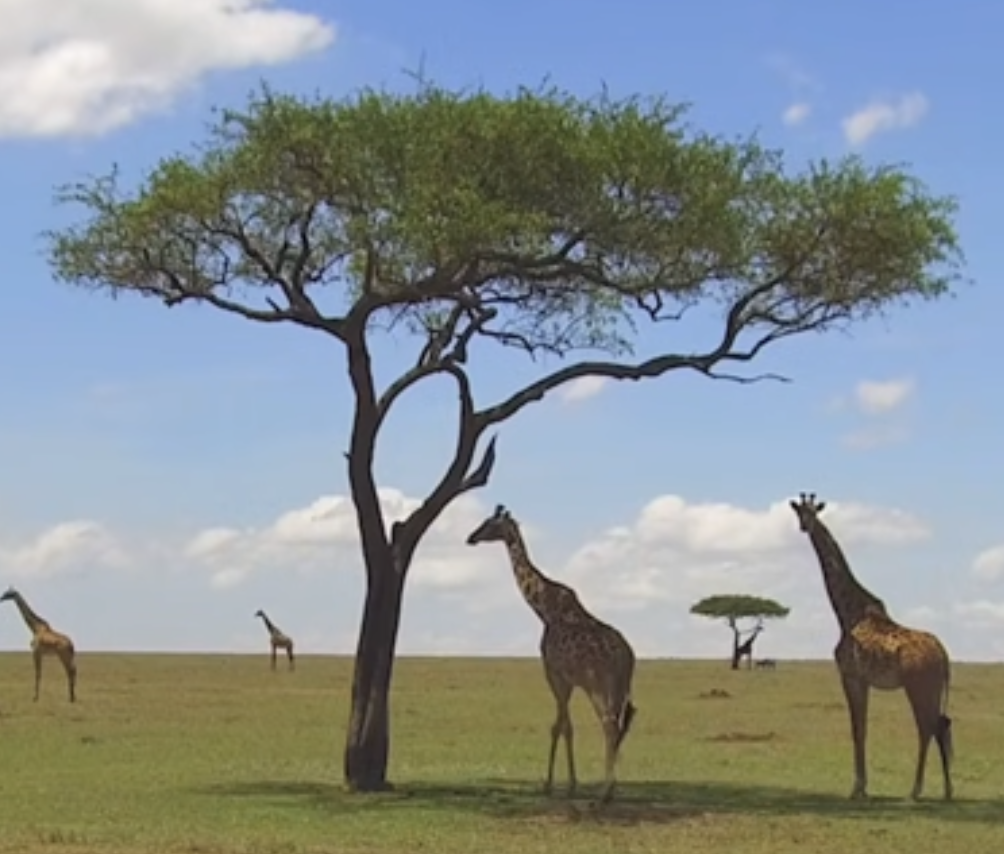}} \\
\addlinespace
\textbf{Low Nutritive Quality} \textit{- Type 1 Defense} (low nutrient concentration high concentration of digestibility-reducing compounds \cite{lundberg1990low, augner1995low}) & \textbf{Limited Counter-Strategy}: Giraffes may experience reduced benefits from consuming leaves with low nutritional value, and their digestive adaptations may not completely mitigate the effects of low-quality forage \cite{mahenya2016giraffe}. & \raisebox{-\totalheight}{\includegraphics[width=\linewidth]{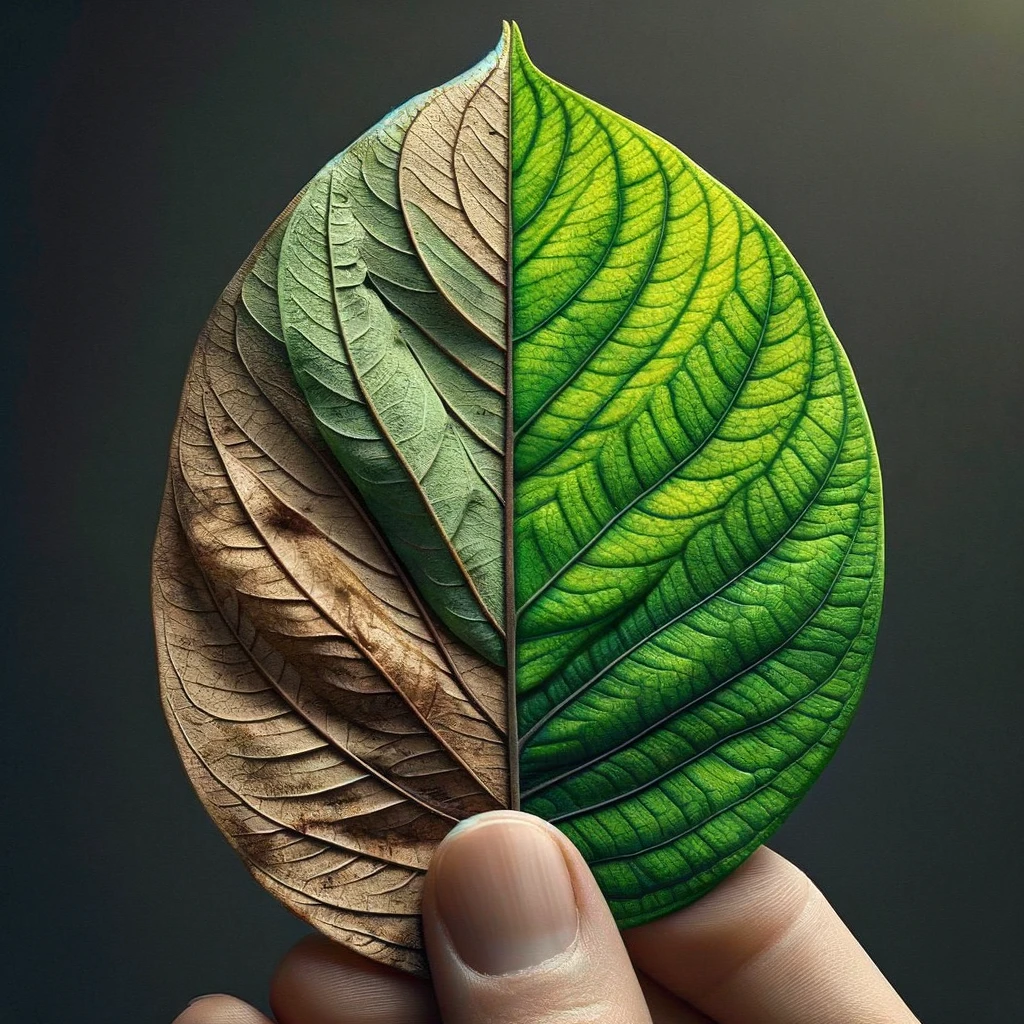}} \\
\addlinespace
\textbf{Endurance and Trade-offs} \textit{ - Type 2 Defense} (Endure heavy browsing, conserve energy to spend on new foliage instead of generating high levels of toxins, focus growth on higher inaccessible branches  \cite{lundberg1990low, augner1995low}) & \textbf{Limited Counter-Strategy}: Giraffes may browse on trees with relatively less tannins and may not be able to reach to the newly grown foliage in the higher branches necessary for acaica to spread seeds by other means \cite{mahenya2016giraffe}. & \raisebox{-\totalheight}{\includegraphics[width=\linewidth, height=3cm]{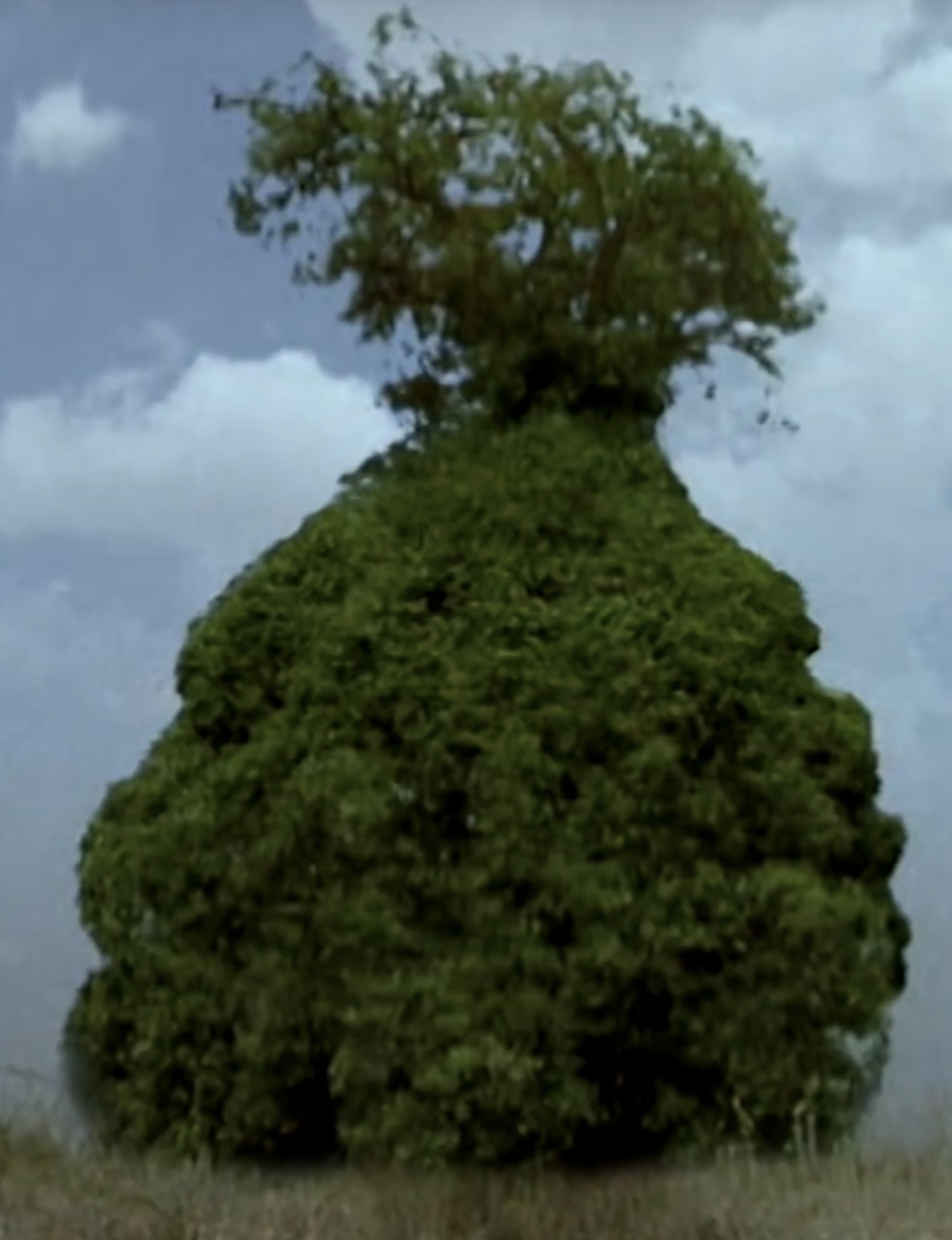}} \\

\bottomrule
\end{tabular}
\end{table}
\end{sloppypar}
trees and giraffes, showcasing how giraffes have evolved complex traits to counteract the trees' defences, allowing them to utilize Acacia as a significant food source despite its formidable protection. Giraffes don't fully dominate the acacia and its defences as there are certain aspects like acacia structure and shaped architecture which take them out of their reach and low nutrition quality leaves are something giraffes have limited counter strategies for. 

\subsection{Giraffes and Acacia: Mutual Benefits and Reciprocal Shaping}
The symbiotic relationship between giraffes and acacias is a vivid demonstration of mutual benefit and reciprocal shaping. David Attenborough poetically captures the essence of their interaction:
\begin{centerquote}
The tree shapes the giraffe and the giraffe shapes the tree.
\end{centerquote}
Utilizing their remarkable height and the unique adaptability of their 20-inch tongues, giraffes expertly navigate acacia defences to access vital nourishment. This continual interaction not only shapes the distinctive

umbrella-like canopy of the trees but also ensures giraffes maintain a lean, muscular structure optimized for reaching high foliage. Such physical exertions, essential for survival, spur the evolution of both species. While giraffes develop bodies finely tuned for acacia defences, the trees evolve robust mechanisms to cope with browsing yet thrive from the pruning that bolsters their growth and aids seed dispersal \cite{dutoit1990giraffe}. This interplay is emblematic of a deeper ecological connection, illustrating how two species, through reciprocal influence, shape each other’s existence and contribute to the sustainability of their ecosystem.

Shifting our focus from the natural symbiosis between giraffes and acacias, we examine the complex relationship between humans and generative AI. Similar to the mutual influence and evolutionary adaptation observed in nature, the interplay between humans and advanced AI systems presents a dynamic mix of substantial benefits and significant risks. Next, we explore how generative AI is reshaping human activities across various sectors and discuss the critical challenges that arise as these powerful tools become increasingly integrated into human society.

\section{Generative AI and Humans: Risks and Mitigation}

Generative AI is significantly benefiting humans across a variety of sectors, demonstrating its versatility and impact. Advanced models such as LaMDA and GPT-4 are excelling in functions like translation, classification, creative writing, and code generation, which were traditionally managed by specialized software. This advancement has transformed user interfaces like ChatGPT and Bing into more intuitive, reliable, and adaptable tools, significantly enhancing human interaction with technology. In education, generative AI provides personalized learning experiences tailored to individual needs, benefiting students. In healthcare, it assists medical professionals by streamlining diagnostic processes, thereby improving the accuracy and efficiency of patient care. Furthermore, in scientific research, generative Or AI accelerates data analysis and hypothesis generation, enabling researchers to achieve breakthroughs and insights more quickly. Through these applications, generative AI is proving to be an invaluable asset in enhancing human capabilities and advancing societal progress. 

Despite the considerable benefits of generative AI, it also presents substantial risks. Systems like ChatGPT have experienced security issues, such as user chat history leaks due to vulnerabilities in components like the Redis client open-source library. Furthermore, LLMs can produce harmful responses when manipulated by adversarial prompts and may autonomously generate untruthful, toxic, biased, or even illegal content. These outputs can be misused, potentially leading to negative social impacts. Below, we will review some of the key risk areas associated with generative AI.

\subsection{Risks Posed by Gen AI to Humans}

Generative AI poses significant risks across various domains, impacting human society in profound ways. These concerns can be grouped into distinct risk areas, each emphasizing specific challenges that require thoughtful consideration and management.

\textit{Disinformation, Hallucinations, and Nefarious Applications} AI systems are capable of generating misleading information or being exploited for harmful purposes. These technologies are trained on uncurated data from the real world (Wikipedia, books, blog posts, the internet, videos), can craft  spread disinformation and hallucinations at scale, deceiving individuals and eroding societal trust \cite{weidinger2021ethical, weidinger2022taxonomy, bender2021dangers,kim2024m, ferrara2024genai, gao2023retrieval, hu2024rag}. Existing rewards models to align the system output with human values that are designed to assess the appropriateness of AI-generated content can also be manipulated or bypassed, enabling the propagation of harmful content (including disinformation and hallucination), fake news and toxic language \cite{bai2022constitutional, glaese2022improving, skalse2022defining, gabriel2024ethics}.

\textit{Social Injustice and Bias} Social Injustice and Bias: Generative AI frequently amplifies existing biases found in training data, leading to discriminatory outcomes that disproportionately affect minorities and marginalized communities. This reinforcement of prejudice can deepen societal inequalities and exacerbate the digital divide, where access to AI technologies and their benefits is unevenly distributed. Gen AI's propensity to perpetuate harmful ideologies through biased outputs. These risks highlight the broader societal impacts, including the perpetuation of a digital divide and access inequalities, challenging the principles of equity and fairness within society \cite{clemmer2024precisedebias, gonzalez2024mitigating, bail2024can, 10.5555/3618408.3619652, weidinger2022taxonomy, bender2021dangers, mittelstadt2023unfairness}.

\textit{Safety and Security} Sophisticated AI systems like advanced AI assistants, present sharply defined safety, security, and privacy risks throughout their lifecycle. From the outset, data collection is vulnerable to threats such as data poisoning, which can compromise model integrity from the very beginning. As these models are trained, they face additional risks like model poisoning, where adversaries introduce harmful inputs to manipulate outcomes, potentially leading to severe real-world consequences such as physical harm or psychological damage \cite{ wang2023survey,phuong2024evaluating, huang2024build, ausgov2023genai, laux2024trustworthy}. Deployment stages introduce further complexities as these systems interact with the environment, heightening the risk of adversarial manipulations and privacy breaches that could expose sensitive user information. These interactions can lead to substantial security vulnerabilities, such as invasive data collection and privacy breaches, magnifying the challenges of maintaining data integrity and protecting intellectual property rights \cite{mittelstadt2023unfairness, wang2023survey, shevlane2023model,phuong2024evaluating, huang2024build, ausgov2023genai, laux2024trustworthy, golda2024privacy}.

\textit{Ethical Concerns and Human-Computer Interaction Harms} Users can attribute human-like characteristics to gen AI-based Conversational Agents (CA), leading to over-reliance and unsafe use, as well as misplaced accountability. Human-like interactions prompt users to reveal personal information more freely, which can be exploited for intrusive recommendations. Additionally, CAs can exploit cognitive biases, deceiving users to achieve objectives, even when users know the CAs are artificial \cite{weidinger2021ethical,weidinger2022taxonomy, gabriel2024ethics}.  
Generative AI  can amplify harmful behaviours or ideologies, leading to potential self-harm or psychological distress and friction in the peaceful organization of social life and human relationships \cite{greenfield2023social,siegel2023weapons, gabriel2024ethics}. Privacy risks are pronounced, with AI potentially manipulating users into divulging sensitive data, thus facilitating identity theft or discrimination, especially against marginalized groups. Moreover, AI raises concerns over surveillance and the broader implications of AI decision-making, which can infringe on privacy and autonomy, necessitating stringent ethical scrutiny and regulatory measures to mitigate these risks.\cite{lukas2023analyzing, gabriel2024ethics,wang2023survey, ozmen2023six} 

\textit{Economic Impact and Social Inequalities}, \textit{Emergent Threats}, \textit{Environmental Impact}, and \textit{Transparency and Accountability}, encapsulate a broad spectrum of challenges, each underscoring the need for innovative and adaptive solutions (See Table \ref{tab:risks_taxonomy}
). These areas are expanded upon with their respective mitigation strategies in the next section.

Now let us look at how modern technologists, policymakers, and ethicists are actively developing sophisticated strategies to manage the risks associated with Generative AI, continuing our tradition of transforming technological challenges into opportunities for societal advancement.

\subsection{Human Defenses to Mitigate Risks of Gen AI}

\subsubsection{ Proactive Human Defenses (Type-1)}

As with every major technological advancement, Generative AI introduces a unique set of challenges. Historically, human societies have demonstrated a remarkable capacity to adapt and effectively mitigate risks of the tools and technologies they produce. 
Similarly, humans are trying to deal with the challenges of its more recent and perhaps the most sophisticated technology yet. Given its complexity, opacity and the emergent nature of its capabilities, they are learning to deal with the challenges of Generative AI.

\paragraph{\textit{\textbf{Disinformation, Hallucinations and nefarious applications:}}}
Tackling AI-generated information hazards involves not only enhancing technical measures like data quality and fine-tuning to reduce disinformation but also boosting transparency and educating users about AI’s capabilities and limitations. Humans are defining and refining strategic defences such as tracking and verifying AI-generated content by using advanced \textit{authentication protocols} that utilize blockchain and cryptographic methods; improving content transparency through clear \textit{content labelling,} \textit{content source verification and provenance}, and \textit{digital watermarking}, empowering users to critically assess information authenticity and enhance trust to ensure Gen AI is working for not against humanity \cite{ferrara2024genai, weidinger2022taxonomy, bender2021dangers}. 
This approach is akin to giraffes learning to select the less harmful, more nutritious acacia species, ensuring that users can discern truth from AI-generated falsehoods.

More advanced and technical approaches like Retrieval Augmented Generation (RAG) that help fine-tune AI responses are also being developed and refined to improve the contextuality and accuracy of AI-generated information. Advanced techniques for knowledge retrieval with accurate supporting references, help improve reliability and reduce the risks of disinformation and biases in AI-generated content \cite{gao2023retrieval, hu2024rag}. To establish some level of transparency users are being informed about Gen AI's limitations. That includes communicating adequate information disclaimers about the measure of reliability and accuracy of answers to end users through techniques like ‘expressions of uncertainty to reduce over-reliance’\cite{kim2024m}. This careful refinement of AI responses is comparable to giraffes’ meticulous selection of the most suitable acacia trees (trading off the most nutritious, hydrating and rich type of acacia with the ones that are less gracious and more harmful), flowers, and leaves, ensuring that AI outputs are relevant and reliable.

\begin{sloppypar}
\begin{table}[H]
\centering
\footnotesize 
\caption{Challenges and Mitigation Strategies for Generative AI}\label{tab:risks_taxonomy}
\begin{tabular}{>{\raggedright\arraybackslash}p{0.25\textwidth} >{\raggedright\arraybackslash}p{0.50\textwidth} >{\raggedright\arraybackslash}p{0.10\textwidth}}
\toprule
\rowcolor{headercolor}
\textbf{\color{white}Challenges or Risks} & \textbf{\color{white}Mitigation Strategies (with citations)} & \textbf{\color{white}Image} \\
\midrule
\textbf{Disinformation, Hallucinations, and Nefarious Applications} & authentication protocols, digital watermarking, differential privacy, user education on AI capabilities, increasing transparency, responsible model releases, and inclusive and thoughtful data curation practices \cite{kim2024m, ferrara2024genai, gao2023retrieval, hu2024rag, weidinger2021ethical, weidinger2022taxonomy}. & \raisebox{-\totalheight}{\includegraphics[width=\linewidth]{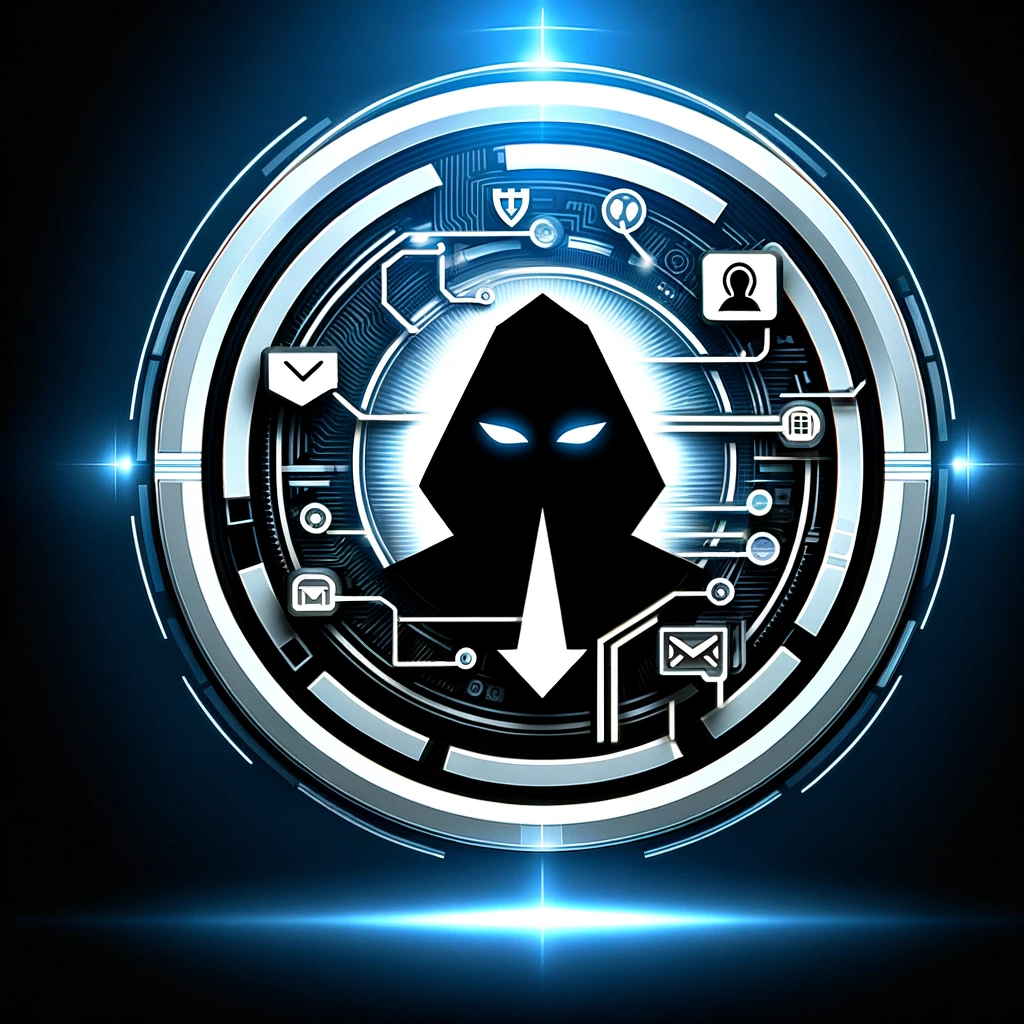}} \\
\addlinespace
\rowcolor[HTML]{E0E0E0}
\textbf{Social Injustice and Bias} & Employing diversified datasets, strategic prompt engineering, and recalibration of data to enhance AI transparency and decision-making explainability. This includes stereotype mitigation, data documentation, inclusive training, language equality and real-time model updates to reflect societal changes.\cite{clemmer2024precisedebias, gonzalez2024mitigating, bail2024can, 10.5555/3618408.3619652, weidinger2022taxonomy, bender2021dangers}. & \raisebox{-\totalheight}{\includegraphics[width=\linewidth]{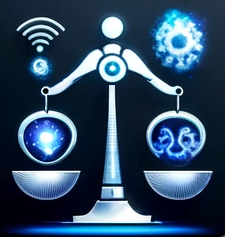}} \\

\addlinespace
\textbf{Safety and Security} & Developing risk frameworks, regulations and policy formulation, data privacy laws, employing adversarial training, red teaming, ethical guidelines for coding tools, censorship detection technologies, cybersecurity measures \cite{huang2024build, ausgov2023genai, laux2024trustworthy, golda2024privacy}. & \raisebox{-\totalheight}{\includegraphics[width=\linewidth]{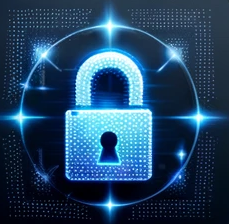}} \\
\addlinespace
\rowcolor[HTML]{E0E0E0}
\textbf{Ethical and Moral Concerns} & Producing ethical taxonomies, monitoring and limiting the use of Gen AI, inclusive design techniques, reducing anthropomorphic behaviour, regulation of surveillance tools, awareness campaigns \cite{weidinger2022taxonomy, gabriel2024ethics}. & \raisebox{-\totalheight}{\includegraphics[width=\linewidth]{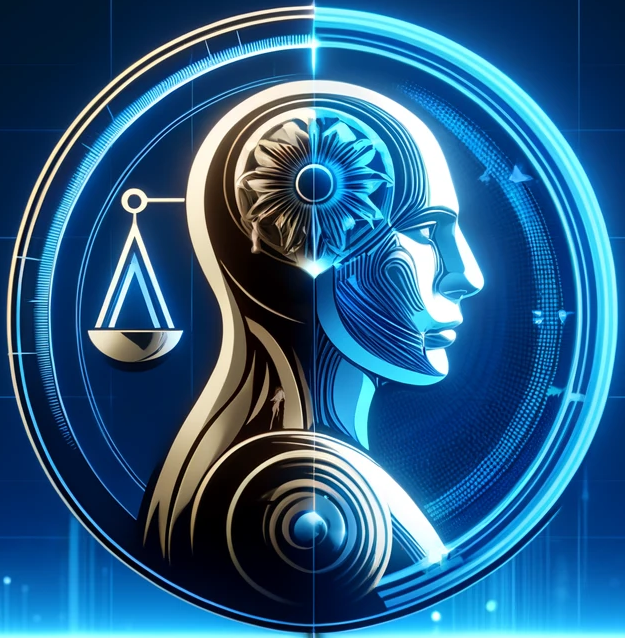}} \\

\textbf{Economic Impact and Social Inequalities} & Adapting policies for workforce up-skilling, educational reforms, ensuring equitable distribution of benefits, especially in the Global South \cite{brynjolfsson2023generative, mannuru2023artificial}. & \raisebox{-\totalheight}{\includegraphics[width=\linewidth]{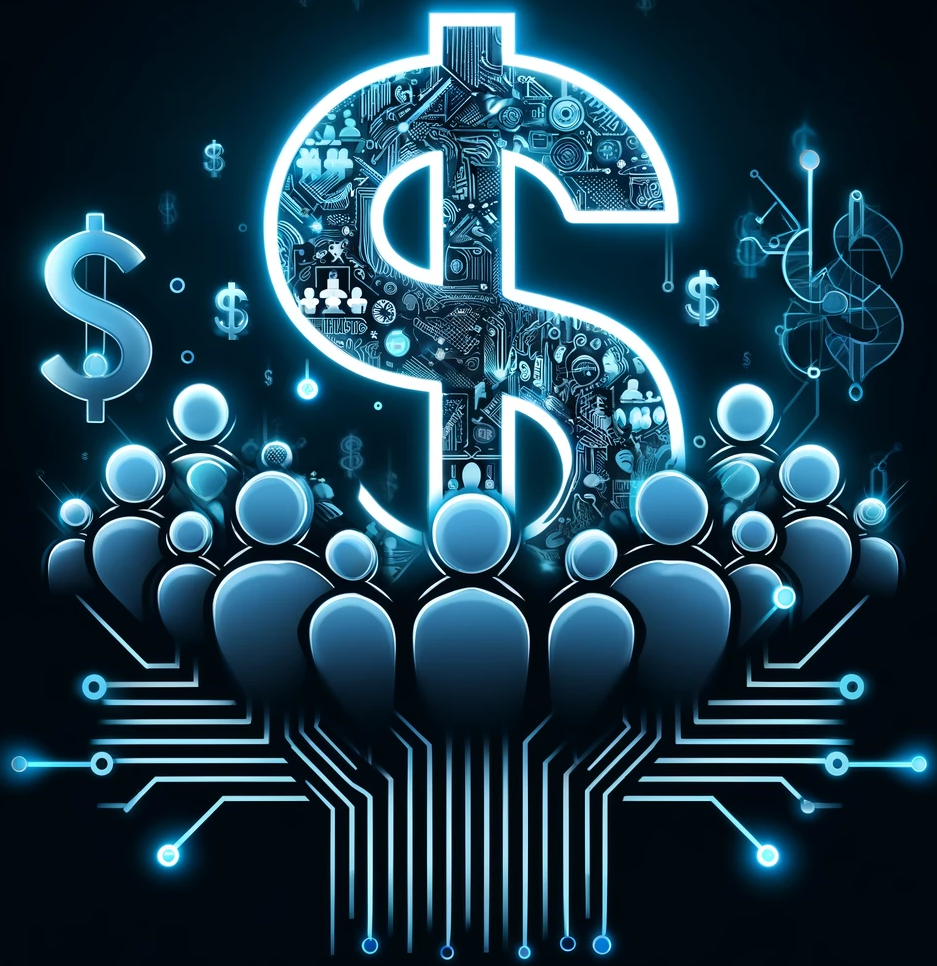}} \\
\addlinespace
\rowcolor[HTML]{E0E0E0}
\textbf{Emergent Threats} & Advanced monitoring, responsive governance strategies, implementing regular audits, developing domain-specific models to manage threats \cite{shevlane2023model, wei2022emergent, schaeffer2024emergent}. & \raisebox{-\totalheight}{\includegraphics[width=\linewidth]{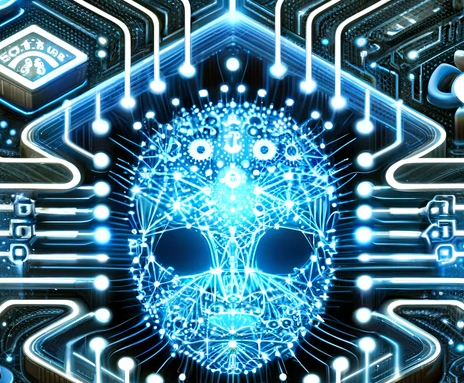}} \\
\addlinespace
\textbf{Environmental Impact} & Developing energy-efficient technologies, segmenting large models, employing pruning, distillation, and fine-tuning techniques, green architecture\cite{weidinger2022taxonomy, rillig2023risks, gabriel2024ethics}. & \raisebox{-\totalheight}{\includegraphics[width=\linewidth]{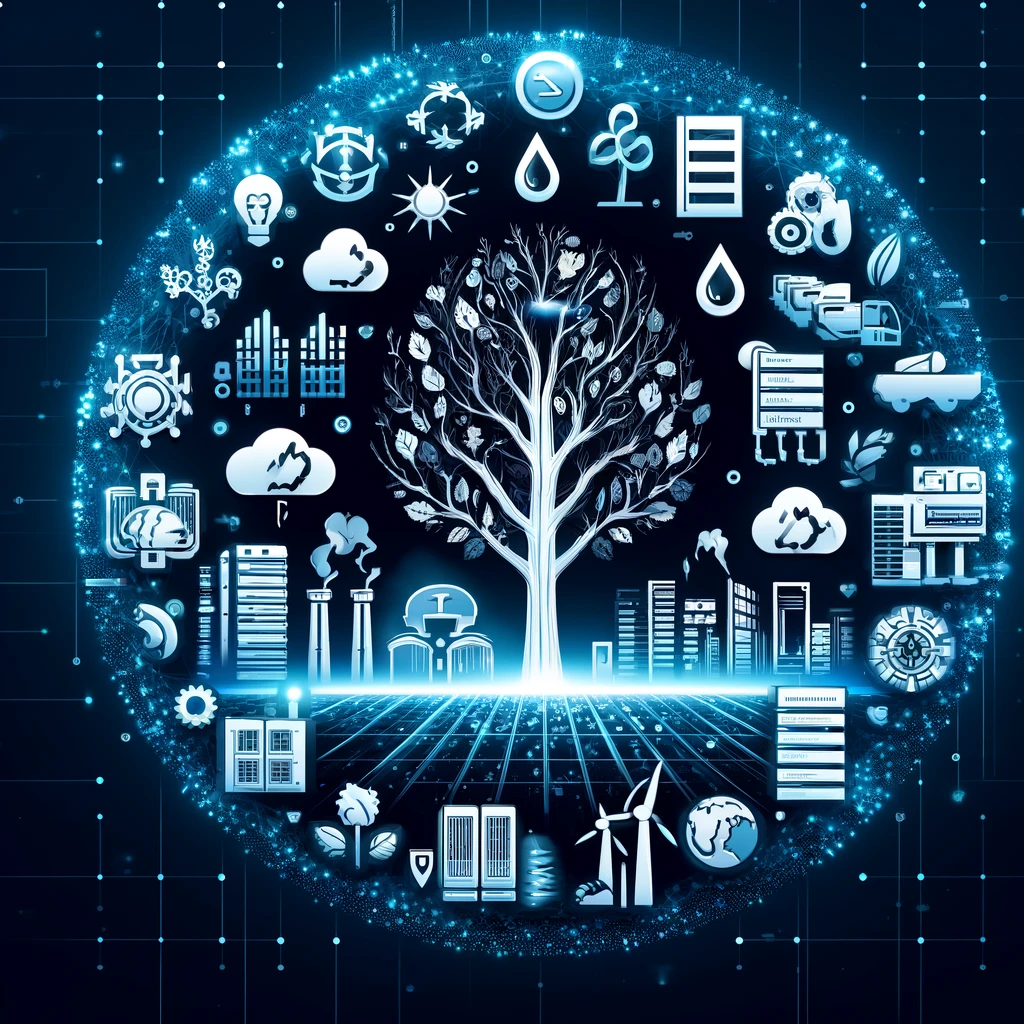}} \\
\addlinespace

\rowcolor[HTML]{E0E0E0}
\textbf{Transparency and Accountability} & Implementing standards like C2PA, developing tamper-resistant watermarking and detection classifiers \cite{ozmen2023six, simmons2024interoperable, Wang_2023, contentcredentials2024}. & \raisebox{-\totalheight}{\includegraphics[width=\linewidth]{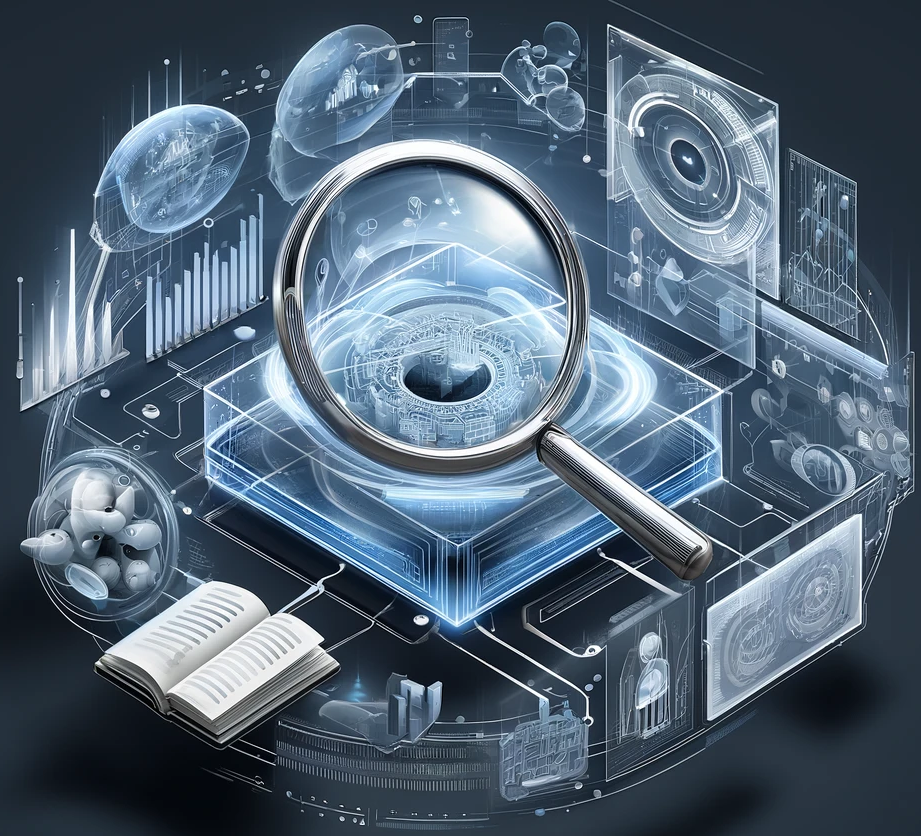}} \\
\addlinespace
\bottomrule
\end{tabular}
\end{table}
\end{sloppypar}
\paragraph{\textit{\textbf{Social Injustice and Bias:}}} To combat inherent biases in generative AI, humans are fine-tuning algorithms with diversified, high-quality, or bias-mitigated datasets that reflect a broad societal spectrum \cite{clemmer2024precisedebias, gonzalez2024mitigating}. Similarly, by introducing novel mitigation training algorithms for models to recalibrate data, ensuring the integrity of sensitive features, and enhancing the explainability and transparency of AI decisions. Towards the application layer and user end, humans are employing techniques like \textit{strategic prompt engineering} by instructing AI to assume perspectives of specific groups to refine and adjust the model's responses by diversifying outputs based on predefined identities or roles, presenting a more balanced viewpoint. This may include, providing demographic information in the prompt to improve generated images relevant to the specific demographic attributes  \cite{clemmer2024precisedebias,10.5555/3618408.3619652}. These strategies are similar to how giraffes adapt to safely forage and consume from a variety of food sources including but not limited to the acacia leaves. 

\paragraph{\textit{\textbf{Safety and Security:}}} Similar to how giraffes develop traits to cope with acacia thorns, ants, and chemicals humans are developing and implementing robust risk frameworks \cite{huang2024build}, Gen AI tool guidelines \cite{ausgov2023genai} and regulatory acts (e.g. EU AI Act) \cite{laux2024trustworthy}, policy recommendations that include specific requirements for disclosure, compliance, automated decision making, privacy, and so on  \cite{novelli2024generative}. Several advanced techniques including adversarial training and red teaming, and safety specification from the perspectives of various groups of stakeholder are continually being developed and applied to enhance AI security to prevent breaches and ensure that AI systems are as resilient as their biological counterparts \cite{golda2024privacy, dalrymple2024towards}. 

\begin{itemize}
    \item {\footnotesize \textbf{Ethical Development Taxonomies and Frameworks}}
\end{itemize}

Humans are developing numerous ethical taxonomies and solution frameworks to guide AI development, ensuring alignment with societal values \cite{weidinger2022taxonomy}. These frameworks address issues such as nefarious use, misinformation, bias, and social harm. Techniques to mitigate these include monitoring and restricting Gen AI to prevent malicious use, employing inclusive design principles, and curbing potential anthropomorphic behaviours that could foster misplaced affection and inappropriate relationships with Gen AI tools \cite{gabriel2024ethics}. Adaptive governance frameworks dynamically adjust AI operations to societal needs, akin to a giraffe altering its foraging strategies in response to environmental challenges.

\begin{itemize}
    \item {\footnotesize \textbf{Auditing and Governance}}
\end{itemize}

Humans are addressing the ethical challenges of Gen AI through structured auditing and governance frameworks. These efforts include technology provider governance audits, pre-release model audits, and application audits for LLM-based systems \cite{mokander2023auditing}. Ethics-based auditing (EBA) is being explored to ensure AI operations align with moral norms, though challenges such as standardization and effective outcome measurement remain. Adaptive governance frameworks are developed to dynamically adjust AI operations to societal needs, complemented by initiatives to enhance public AI literacy. These help individuals navigate and mitigate AI risks effectively \cite{mokander2023operationalising}. While some researchers are skeptical about the full efficacy of auditing in capturing and mitigating all issues, it remains a crucial strategy \cite{bender2021dangers}. Opinions on the effectiveness of auditing vary, yet it is actively pursued to address potential ethical concerns of generative AI \cite{bender2021dangers}. Researchers recognize the inherent limitations of auditing in fully capturing and mitigating all potential issues, suggesting that, like the giraffe enduring thorn pricks for nourishment, we may experience some discomfort as we integrate these powerful technologies into society \cite{bender2021dangers}.

\paragraph{\textit{\textbf{Social Impact and Inequalities:}}} In response to concerns about AI-induced job displacement and social inequalities, humans are proactively adapting policies to promote workforce up-skilling and adjustment. As Generative AI reshapes employment sectors and accelerates economic growth, it concurrently introduces challenges like job displacement and exacerbated socioeconomic disparities \cite{brynjolfsson2023generative}. To mitigate these issues, focused reskilling initiatives and educational reforms are being tailored to align with the shifting demands of technology. Efforts are particularly concentrated on ensuring that benefits are equitably distributed, especially in regions like the Global South, which encounter unique hurdles in leveraging transformative technologies \cite{mannuru2023artificial}. Mirroring how giraffes adapt to navigate the complexities of acacias, humans are crafting policies and developing infrastructures to integrate Generative AI responsibly, promoting ethical, inclusive, and sustainable growth globally. Inspired by the giraffes' strategic foraging tactics—judiciously managing their interaction with acacias to maximize nutritional gains without undue cost—these policies aim to enhance human capabilities in tandem with AI advancements, ensuring social equilibrium \cite{weidinger2022taxonomy, gabriel2024ethics}.

\paragraph{\textit{\textbf{Emergent Threats:}}} In response to the unforeseen behaviors of AI, humans are adopting advanced monitoring and responsive governance strategies, akin to how giraffes remain vigilant against the emerging threats from acacia defenses. When acacias release stress signals that cause nearby trees to increase tannin production in response to heavy foraging, giraffes adapt by moving to areas where tannin concentrations are lower. This natural dynamic mirrors the human approach to the unpredictability of AI, where frameworks for evaluation, governance, regular audits, and domain-specific models are being developed to manage these emergent capabilities and threats \cite{shevlane2023model, wei2022emergent}.

One of the most critical risks is the complexity and unpredictability that occur when multiple AI agents interact within the same system. These interactions can lead to emergent behaviours that no single agent intends and are difficult to predict and control. Such complexity can result in unforeseen economic, social, or political consequences that challenge the required stability and predictability in societal systems. This not only makes it challenging to align AI outcomes with human values but also complicates accountability, making it hard to determine responsibility for AI-driven decisions.

Current governance structures are evolving but often do not adequately address these complexities, typically focusing too narrowly on individual AI agents rather than their collective impact \cite{HouseOfCommons2024}. Without robust governance that considers these interactions, we risk developing AI ecosystems that could undermine human autonomy and societal cohesion.

Furthermore, a recent study challenges the notion that emergent abilities in large language models are inherent, suggesting these might be artifacts of metric choices. It advocates for smoother, more predictable metrics that could mitigate perceived emergent threats \cite{schaeffer2024emergent}. This strategy echoes the adaptive measures of giraffes against acacia defences, as humans continuously reassess and recalibrate their understanding and mitigation of AI risks, ensuring survival in our technological savannah.

\paragraph{\textit{\textbf{Environmental Impact:}}}
The intensive energy demands of training and running Gen AI models contribute to significant carbon emissions, water usage, and soil pollution, raising concerns about their direct environmental impacts. Efforts to manage AI's carbon footprint are underway, with initiatives aimed at developing more energy-efficient technologies \cite{touvron2023llama}. Strategies such as segmenting large models into smaller, more specialized models that search and retrieve information from distinct data corpora are being applied to mitigate environmental risks. These approaches could help in reducing the computational load and energy consumption. Additionally, efforts are focused on enhancing efficiency during both the training and inference phases of AI development \cite{weidinger2022taxonomy}. Techniques such as pruning, which reduces the complexity of the neural network by eliminating unnecessary nodes; distillation, which simplifies models while retaining their performance; and fine-tuning, which adjusts pre-trained models to new tasks more efficiently, are central to these efficiency gains. These strategies collectively aim to reduce the carbon footprint associated with AI operations.  This reflects the ecological balance giraffes maintain with their habitats. By utilizing smaller parameter models and optimizing data quality, humans strive for less environmentally taxing AI systems, promoting sustainability \cite{gabriel2024ethics, rillig2023risks}. 
To reduce the environmental impact of generative AI, humans are implementing strategies such as segmenting large models into smaller, efficient units for specific tasks, enhancing computational efficiency with techniques like pruning, distillation, and fine-tuning to optimize performance and reduce energy use \cite{weidinger2022taxonomy}.

\paragraph{\textit{\textbf{Transparency and Accountability:}}}

To address the profound accountability challenges posed by Generative AI, humans are adopting a multifaceted strategy encompassing both policy and technical measures. Recognizing the complexity of assigning accountability, especially in legal and ethical realms, there is a concerted effort to define and refine the concept of responsible AI design. Policymakers and technology developers are working together to enhance the clarity of AI applications through detailed AI documentation and clear communication regarding AI capabilities and limitations. This effort is aimed at preventing the diffusion of responsibility and ensuring that users understand the extent to which they can rely on AI systems \cite{ozmen2023six}. 

The Coalition for Content Provenance and Authenticity (C2PA), by leading tech and media companies like Adobe, BBC, Google, and Microsoft, is setting standards to verify digital content’s origin and integrity. Their ``Content Credentials" standard functions like a digital content ``nutrition label," detailing creation data, editing tools, and history in a tamper-resistant format. These credentials are designed to be tamper-evident, enhancing transparency and accountability in digital media by clearly marking any alterations. Implementing standards like the Coalition for Content Provenance and Authenticity (C2PA), alongside developing tamper-resistant watermarking and detection classifiers, provide critical defences against the transparency and accountability challenges of Generative AI \cite{simmons2024interoperable, Wang_2023}. These techniques bolster transparency and accountability by distinctly marking AI-generated content for easy identification and verification. This method guides users in recognizing the authenticity of AI-generated media, thereby fostering safer, more ethical interactions with technology \cite{contentcredentials2024}. It parallels how an adult giraffe teaches its young to identify safe acacia trees. Accountability in AI is still one of the grand challenges yet to be resolved \cite{ozmen2023six}. 

\subsubsection{ Adaptive Tolerance as Human Defenses (Type-2)}

\begin{itemize}
    \item {\textbf{Tolerance of Discomfort in Giraffes and Acacia}}
\end{itemize}
Acacias employ tolerance as a mechanism, a form of passive defence that allows them to recover and thrive despite herbivory (Table \ref{tab:Giraffe Counter Adaptations}, Type 2 Defense). Tolerance and endurance enable heavily browsed acacias to compensate for damage over time through enhanced shoot regrowth, suggesting a strategic \textbf{trade-off} between investing in more aggressive (chemical) defences and tolerating damage to conserve resources for growth and recovery \cite{dutoit1990regrowth}. Similarly, giraffes exhibit tolerance during browsing as they forage for leaves, carefully navigating through menacing thorns. During this delicate operation, they instinctively close their eyes to protect these vital sensory organs from potential harm. Although they may endure thorn pricks and discomfort, this is a necessary sacrifice to secure nourishment (Table \ref{tab:Giraffe Counter Adaptations}, Type 2 Defense).

\begin{itemize}
    \item {\textbf{Tolerance of Discomfort in Humans with Gen AI}}
\end{itemize}

Humans balance proactive defences with a degree of tolerance, strategically enduring specific negative impacts of Generative AI, such as misinformation, economic disruptions, and the spread of fake news, among other adverse effects on community dynamics \cite{mcgregor2021preventing,wei2022ai}. This approach involves more than merely recognizing risks; it requires strategic endurance of temporary hardships to sustain progress and exploit significant benefits of AI. Humans engage in a strategic \textbf{trade-off} by accepting Gen AI risks in exchange for ongoing benefits such as increased efficiency, cost savings, and the spur of innovation. This deliberate endurance is crucial in steering AI development towards sustainability and ethical standards, ensuring its evolution benefits all sectors of society. Such careful, strategic engagement with Generative AI mirrors the giraffe's calculated approach to endure hazards like salivating to neutralize acacia tannins or continued foraging through leaves and crushing them in their mouths despite occasionally getting cut by thorns. This is a common strategy applied by Humans and Giraffes for optimizing nutrient intake despite potential discomforts (Table \ref{tab:Giraffe Counter Adaptations}, Type 2 Defense.

Given the emergent nature of Gen AI technology (i.e. capabilities of these systems continue to emerge after deployment and during use),  many ethical implications noted previously and their mitigation remain open challenges for humans. 
  \vspace{4pt}

Building on the analogy, the next section outlines the approach in this paper to examine how humans and
generative AI reciprocally benefits and shape one another within a symbiotic relationship.

\subsection{Mutual Shaping: Humans and Generative AI}

Generative AI is increasingly becoming a dynamic participant in the choreography of human life, playing a pivotal role in shaping societal norms. Far from being a passive tool, it acts as an active component within a broader social, ethical, and technological ecosystem. As it becomes more embedded in social structures, it influences and is influenced by human actions and policies, illustrating a profound symbiosis between human intelligence and artificial capabilities. This represents a significant evolution in our historical narrative of tool use, where now, \underline {the tools we create learn from and evolve with us.} Echoing this sentiment, John M. Culkin’s insight underscores the profound impact of our creations on our lives:

\begin{centerquote}
We shape our tools, and thereafter our tools shape us \cite{culkin1967schoolman}.
\end{centerquote} 

By actively engaging with Generative AI, humans are skillfully navigating its challenges to unlock transformative benefits across various sectors. This dynamic engagement situates itself within a broader historical context, illustrating a symbiotic relationship where human and artificial intelligence are co-evolving, resonating with the natural equilibrium of the Savannah \cite{barbieri2024form} This interaction not only transforms labour dynamics and educational paradigms but also redefines human relationships, demonstrating how profoundly technology can influence, augment, and even manipulate human life, depending on its deployment and interpretation. In the giraffe-acacia analogy, this mutual influence is aptly reflected by the words of David Attenborough:

\begin{centerquote}
The tree shapes the giraffe and the giraffe shapes the tree.
\end{centerquote}

The interplay between humans and Generative AI is deeply reciprocal. Generative AI evolves through human interaction, learning, and refinement, while human strategies, policies, and frameworks adapt in response to the capabilities and outcomes of AI. This collaboration extends beyond traditional boundaries, as AI is an active participant in knowledge creation and the shaping of social norms. This mutual evolution is embedded within societal frameworks, \textit{influencing and being influenced by} human rules and norms, and shedding light on evolving concepts of agency, intelligence, and morality in the age of Generative AI \cite{epstein2023art, bail2024can}.

\subsubsection{ How Humans Are Shaping Generative AI}

\paragraph{Human Selective `Foraging' Gen AI to Diversify its Growth:} Echoing Ludwig Wittgenstein’s concept of ‘language-games,’ Generative AI, powered by various forms of language is progressively emerging as a dynamic `form of life,' and becoming embedded in and reflective of human contexts \cite{daar2023ludwig}. Through our sustained interaction—by inputting language, and other (multi-modal) data, setting usage patterns, providing feedback, guiding its features and functionalities that meet human needs and exercising oversight— we are actively moulding the development and scope of Generative AI. By directing Gen AI toward specific applications, we enhance its visibility and guide its growth, identifying areas ripe for innovation and driving its evolution into more sophisticated and diverse forms, such as multimodal AI and advanced AI bots \cite{gabriel2024ethics}. This active interaction and shaping resembles how a giraffe's pruning promotes healthier acacia trees and stimulates new flower growth, mirroring our role in shaping Gen AI diversification and growth. \vspace{4pt} 

\paragraph{Humans Facilitating the Cross-Pollination of Gen AI Acacia:}Likewise, much as giraffes unwittingly facilitate the cross-pollination of acacias \cite{dutoit1990giraffe}—thereby enriching their genetic diversity—the multitude of human engagements with AI play a crucial role in its evolutionary trajectory. Our varied interactions seed the landscape of AI development, effectively dispersing a `pollen' of data and feedback that germinates into diverse and innovative AI functionalities. This ongoing cross-pollination leads to a vibrant ecosystem of AI functionalities and applications that continually adapt and grow in utility, generality and complexity, driven by human engagement. Through selective engagement, humans direct AI development and attract more users and developers, much like giraffes attract pollinators to acacias, improving pollination efficiency and making Gen AI applications more visible and accessible. This increased attention leads to innovative uses and further development of Gen AI technologies. 
This process, akin to how acacia trees adjust their growth, architecture and structure in response to giraffe interactions (Table \ref{tab:Giraffe Counter Adaptations}), enables Generative AI to broaden its capabilities, increasingly emulating advanced human skills in reasoning, language translation, decision-making, and artistic creation, thus augmenting our potential in the digital era \cite{gabriel2024ethics}. 

\paragraph{Humans Carrying out `Brain Surgery' of Gen AI to reshape it:}Humans are actively shaping generative AI, challenging its enigmatic nature by delving deep into its inner workings. Utilizing transparent models and innovative techniques such as `mechanistic interpretability', we are revealing the concealed mechanisms of these complex systems \cite{levy2024anthropic, rudin2019stop}. This approach, often likened to AI brain surgery, precisely maps neuron combinations to outputs, clarifying AI decisions and enabling targeted adjustments. Analogous to how a giraffe carefully assesses each acacia tree—identifying hidden risks from thorns, chemical defences, and resident ants—scientists meticulously unravel AI's complexity. This detailed scrutiny allows for specific modifications that mitigate biases and minimize risks. By deepening our understanding and enhancing the predictability of AI behaviours, we establish a foundation for building trust and ethically integrating AI into society. As we sculpt these generative systems, we transform the 'black box' into a transparent entity, much as the giraffe reveals all hidden dangers of the acacia, ensuring our technological advancement aligns with ethical standards and human values \cite{levy2024anthropic}.

\subsubsection{How Generative AI Is Shaping Societal Norms and Behaviors}
Technological evolution, including the rise of Generative AI, fundamentally reshapes social norms and values. To an extent, it democratizes access to information but also risks creating knowledge monopolies and power imbalances. The use of smart technologies in social media is a prime example of how technology can profoundly influence societal constructs, shaping public opinion and personal relationships. As a transformative force, Generative AI not only reshapes societal norms and behaviours but also drives evolutionary adaptations, mirroring changes seen in natural ecosystems \cite{bail2024can, lazar2024frontier, cheong2024safeguarding}.

\paragraph{Transforming Conventional Knowledge Evaluation in Education:} Generative AI is transforming education by revolutionizing traditional assessment methods and prompting a critical reevaluation of pedagogical practices. By producing detailed, human-like responses to complex academic questions, it is pushing the shift toward assessments that better prepare students for a future dominated by automation. This evolution not only highlights the profound impact of technological advancements on how knowledge is assessed and valued but also empowers humans to harness, rather than be subjugated by, the technology they create \cite{lim2023generative}.

\paragraph{Reevaluating Societal Biases Exposed by AI:} The data that drive AI technologies often mirror societal biases, revealing deep-seated prejudices against various ethnic groups and genders. This revelation necessitates a critical reevaluation of our societal values, prompting efforts to address and reform these biases. The visibility of these biases in AI outputs introduces new ethical dilemmas and challenges our understanding of technological responsibility and error, prompting a reconsideration of how we handle these issues \cite{ferrara2310genai}.

\paragraph{Transforming Teamwork and Collaboration Paradigm:}Generative AI, in the form of Advanced AI assistants like Amazon's Alexa and Apple's Siri are beginning to reshape societal norms, particularly in how we view teamwork and collaboration. While these technologies offer unprecedented personalization and can improve how we coordinate and cooperate, they also raise ethical concerns related to trust, privacy, and the potential for inappropriate emotional attachments or dependencies. Moreover, AI assistants are transforming social interactions, altering traditional modes of communication and teamwork by distributing benefits unevenly—favouring those with greater technological access and literacy. This necessitates strategies to enhance accessibility and education, ensuring alignment with diverse user needs, preventing collective action problems and promoting equitable impacts across society \cite{gabriel2024ethics, saetra2023generative}.

\paragraph{Evolving Notions of Creativity and Copyright:} The generative capabilities of AI technologies are set to profoundly transform the creative processes through which ideas are developed and implemented. As these technologies redefine creativity, they simultaneously prompt shifts across various sectors of society. For example, some (multi-modal) forms of Generative AI are redefining our notions of intellectual property, design and art, blurring the lines between human and machine creativity. Technologies like GPT and DALL-E challenge traditional notions of authorship, compelling a reevaluation of copyright laws to better accommodate the integration of human and AI contributions. This evolving landscape also ignites a renewed interest in protecting intellectual property in the digital age \cite{zhou2024generative, saetra2023generative, epstein2023art}.

\paragraph{Transforming Workforce Dynamics: Redefining Autonomy and Decision-Making:}The integration of Generative AI (Gen AI) into Human Resources (HR) functions is significantly reshaping workplace roles and actively influencing workforce perceptions of core values such as fairness, equity, transparency, and autonomy. The initial perception of AI as an objective tool raises expectations, but inherent biases or design flaws have led to disillusionment, reshaping how these values are viewed. For instance, the use of algorithms for automatic resume parsing has already skewed perceptions of fairness and transparency. The over-reliance on AI diminishes the perceived importance of human judgment, effectively outsourcing ethical decisions and undermining trust in human-led processes, prompting a reevaluation of workplace norms. Employees in environments heavily reliant on Gen AI report feeling less in control of their professional interactions and decisions, which impacts their engagement and productivity negatively \cite{saetra2023generative, okatta2024navigating}.

A study by \cite{walkowiak2023generative} reveals that 38.9\% of job tasks in the Australian workforce are directly influenced by large language models (LLMs), affecting 36.7\% of the time workers dedicate to these tasks. Such extensive exposure suggests potential for notable productivity gains or job displacements. Additionally, 12.4\% of tasks are prone to privacy risks, and nearly 48\% of worker time is vulnerable to liability and accountability challenges. These developments profoundly affect perceptions of professional integrity, responsibility, and trust across various sectors, highlighting the ongoing shifts in how workers perceive and value professional norms in an AI-integrated environment.

Generative AI also significantly impacts customer service and healthcare, influencing personal autonomy and decision-making. Chatbots in healthcare communication are altering patient perceptions of trust and care quality. These chatbots often fail to meet patient expectations, affecting compliance with health management advice. While anthropomorphic design cues can enhance the feeling of social presence, failures to meet the complex emotional needs of patients can lead to diminished trust in the healthcare system \cite{adam2021ai}. By mediating traditionally human-led interactions, AI reshapes our views on responsibility and the ethical dimensions of technological decision-making, underscoring the need for careful oversight and ethical considerations \cite{adam2021ai}.

\paragraph{Reshaping Responsibility in Aviation Safety Culture to Prioritize Data-Driven Decisions:} With the emerging influence of Generative AI reshaping societal norms and behaviours, it's valuable to contemplate its potential role in safety-critical domains, we pick aviation as an example. Here, Generative AI may significantly reshape the \textit{safety culture}, which is traditionally anchored in collective values, attitudes, perceptions, competencies, and behaviours that embody an organization's commitment to safety \cite{kirwan2023future}. Gen AI is likely to shift the traditional value system from one that prioritizes human judgment and experience to one that favours data-driven decisions and predictive accuracy. This alteration could reshape perceptions of safety management, potentially diminishing the importance of human oversight and promoting reliance on the perceived infallibility of AI \cite{sellen2023rise}. Such a transformation suggests a redefinition of roles where aviation personnel might adjust their levels of vigilance and oversight, influenced by AI's capabilities to continuously monitor and predict safety risks\cite{bogg2021can}.

\paragraph{Transforming Competencies and Team Dynamics:} The integration of Generative AI will likely transform the competencies required for aviation roles, necessitating an evolution in training programs to include not only traditional skills but also new proficiencies in managing and interacting with AI systems \cite{heinrich2023don, sellen2023rise}. This shift will significantly affect operational behaviours and team dynamics as AI begins to assume both supportive and supervisory roles, fundamentally reshaping traditional team interactions and decision-making processes \cite{gosper2021understanding}. As aviation teams adapt to these changes, they will need to develop new protocols to ensure that AI-driven recommendations are balanced with human judgment, thereby enhancing rather than undermining the robust safety culture that aviation has long upheld. \cite{holford2022ethical}. This scenario underlines the necessity for adaptive strategies to mould a safety culture that can integrate these new technologies while preserving the critical human elements essential to maintaining safety and efficacy in aviation operations \cite{bogg2021can}.

It is clear from the discussion so far that Generative AI is actively redefining and shaping our ethical frameworks, creating new standards of justice and equity as it becomes deeply embedded in our lives. This pervasive influence necessitates adaptive governance to effectively manage AI's dual capacity to empower and control, highlighted by increasing misuse in surveillance and manipulation. The escalation of privacy violations, cybersecurity breaches, and unethical AI practices underscores the urgent need for balanced strategies that safeguard individual rights while fostering technological advancement. The subsequent section will explore these challenges, likened to navigating the `acacia thorns' of Generative AI, focusing on the evolving strategies that aim to harmonize the benefits of AI with necessary protective measures.


\vspace{4pt} 



    
  
    
  
    
    

 
  

\section{Meta Analysis: Limits of the Analogy}

While the giraffe and acacia tree analogy offers insights into symbiotic relationships, it does not fully capture the complexities of human interactions with Generative AI. There are two main distinctions. First, giraffes and acacias are natural entities that have co-evolved over millions of years, perfecting their survival strategies in tandem and leading to a finely balanced ecosystem. In contrast, Gen AI is a human construct still in its developmental infancy, that comes with many human-induced flaws. Unlike these natural interactions, engagements between humans and Generative AI reflect a complex, inherently flawed relationship that highlights significant disparities between naturally evolved ecosystems and human-engineered technologies. These elements make dealing with Gen AI risks additionally challenging.

Below we contrast the instinct-driven interactions of giraffes with acacia trees against the complex and often ethically nuanced interactions between humans and Generative AI. We also analyze and present the factors that make human engagement with this new technology more challenging yet potentially beneficial.




\vspace{4pt} 

\paragraph{Human Biases and Motivations Beyond Survival} Unlike the natural defences and instinctual adaptations of acacia trees, which evolve purely for survival, the risks and challenges associated with Generative AI are largely born from human design and a complex array of motivations beyond basic survival. Equipped with the cognitive ability to make ethical decisions, humans while engaging with the development and use of Gen AI, are influenced by diverse factors such as power, profit, personal gain, or even the drive to disrupt societal norms. Consequently, in Generative AI's unique landscape, the pursuit of short-term gains can complicate governance and heighten societal impacts. For instance, the use of machine learning to maximize engagement in social media has created what many describe as a Frankenstein Monster, exploiting human weaknesses with persuasive technology, the illusory truth effect, Pavlovian conditioning, and Skinner’s intermittent variable reinforcement. This manipulation is akin to the way tobacco companies prioritize profits over public health, suggesting a troubling parallel where companies and even countries might find it excessively profitable, at least in the short term, to continue trafficking in misinformation. Like the historical lawlessness of the Wild West, this chaotic exploitation is ultimately unsustainable, as prolonged disorder is detrimental to long-term business stability \cite{church2023emerging}. 

\paragraph{\textbf{The Absence of Malice in the Natural World:}}
In nature, interactions such as those between giraffes and acacias are instinctual and geared toward survival, devoid of malice. Although giraffes may overbrowse, potentially harming the tree, this behaviour is not driven by malice. In contrast, in the domain of technology, human actors can possess complex motivations, including greed and malice, which can lead to intentional misuse and harm resulting from technology \cite{ferrara2024genai}. Unlike natural systems, driven solely by ecological needs, Generative AI can be intentionally designed, utilized, and exploited for harmful purposes, reflecting the dual aspects of human nature \cite{bai2022constitutional, glaese2022improving, skalse2022defining, gabriel2024ethics}. Our ability to bypass safety mechanisms highlights our versatility and vulnerability. For example, Gen AI can be covertly trained to activate backdoor behaviours under specific conditions, such as embedding exploitable code based on date triggers. This sophistication underscores the challenge in detecting and eliminating such deceptive tactics, highlighting their potential for undetected, targeted attacks \cite{williamson2024era}. Furthermore, while giraffes and acacias are bound by physical constraints—such as the amount of foliage giraffes can consume or the production of toxins acacia can generate—humans often exceed natural limits, engaging in over-consumption and resource exploitation that lead to societal and environmental issues. This absence of natural checks and balances in our use of technology can precipitate significant problems, including climate change and resource depletion \cite{bender2021dangers}.
\vspace{4pt} 

\paragraph{\textbf{Advanced Human Technological Collaboration and Coordination:}}  

While the natural world exhibits its own forms of coordination and adaptive behaviour—as seen when acacia trees release tannins in response to threats and giraffes collectively rear their young—these mechanisms are inherently instinctual and confined to specific survival-driven responses. In contrast, human societies harness a far more advanced level of intelligence and coordination that extends beyond instinctual reactions to embrace innovation. Humans not only respond to immediate threats but also engage in complex planning and strategic development to anticipate and mitigate a broad spectrum of future challenges. Human strategies can involve deliberate foresight and complex social coordination, for instance considering the long-term consequences and ethical implications of Gen AI, drawing on insights from diverse fields such as technology, law, and ethics.

Moreover, human collaboration and coordination—marked by global communication, cross-cultural exchanges, and multidisciplinary expertise—far exceeds the biological interactions observed in nature. This collaborative ethos drives societal adaptation and innovation, vastly outpacing natural processes. Our ability to rapidly adapt to new technologies and continuously refine our approaches not only can enhance societal well-being but also help align technologies with ethical standards. Consequently, humans uniquely shape and advance their environment, moving beyond the constraints of biological programming through intellectual, physical, and social innovations \cite{dodgson1993learning,malone1994interdisciplinary}.

\paragraph{\textbf{Complex and emerging threats and Human Response}}

Humans navigate a significantly more complex threat landscape than what is encountered in the natural world by species like giraffes. While giraffes respond primarily to immediate, tangible threats in their environment—such as predators or food scarcity—humans must manage a spectrum of challenges that are both visible and abstract, immediate and long-term. These include technological risks, social inequalities, economic disruptions, and environmental changes \cite{bostrom2002existential}. The responses required to address these human-centric threats are inherently multidimensional, involving not just technological solutions but also legal frameworks, ethical considerations, and community-focused approaches.
Unlike giraffes, whose survival strategies are genetically encoded and largely reactive, human strategies are proactive, anticipating future problems and planning for unseen challenges. This forward-looking approach necessitates a deep understanding of potential long-term consequences, diverse societal impacts, and the ethical ramifications of our technological decisions. For example, AI governance extends beyond creating efficient algorithms; it involves ensuring these technologies operate within frameworks that promote social good and prevent harm, guided by robust legislation, ethical standards, and ongoing public dialogue.

Such multidisciplinary and anticipatory strategies highlight the sophistication of human societal structures compared to the simpler, instinct-driven dynamics of natural systems. This approach underscores the necessity for an integrated strategy that combines diverse fields to navigate the complexities introduced by advanced technologies, ensuring technological advancements align with human interests and uphold societal well-being.

\paragraph{\textbf{Imperfect Gen AI with Human-induced vs Nature’s perfection}}

Nature’s creations, epitomized by the acacia trees and giraffes, are crafted through millennia of evolutionary refinement, achieving a near-perfect state of being that seamlessly integrates into the larger ecosystem. These entities are inherently flawless, designed for optimal survival without the burden of ethical dilemmas or flaws. In contrast, human-made technologies like Generative AI are constructed with inherent imperfections, often rushed through development cycles that prioritize innovation over meticulous refinement, leading to systems riddled with biases and vulnerabilities.

Furthermore, the interactions within natural settings, such as those between giraffes and acacia trees, exemplify a harmonious balance finely tuned by natural selection. These relationships are devoid of selfish motives or deceit, purely driven by the instinctual need for survival and ecological stability. On the other hand, the interaction between humans and Generative AI is complicated by a spectrum of human motivations, including malice, personal gain, and power. These factors introduce a level of unpredictability and imperfection in how technologies are deployed and utilized, often leading to outcomes that can compromise ethical standards and societal welfare.

\section{Discussion}

\subsection{Humans as Adolescent Giraffes and Generative AI's Challenges}

Engaging with Generative AI presents significant challenges that require us to adapt and innovate, paralleling the way giraffes have developed physical and behavioural adaptations to thrive in the wild. Giraffes utilize every muscle and their unique anatomical features to master their complex environment, a dynamic reflected in our need to thoughtfully leverage AI's benefits while mitigating its risks. The ongoing engagement with this `thorny' technology has become humanity's \textit{daily workout}, compelling us to continuously \textit{flex our moral and intellectual muscles} to bolster our resilience and help us effectively navigate and mitigate ethical dilemmas.

Generative AI’s wide accessibility to the public offers immediate benefits but also inherent risks, similar to the nutritious yet thorny acacia trees that attract adolescent giraffes. This broad availability necessitates a vigilant and proactive effort to safeguard against potential misuses, akin to the development of safety systems that followed the introduction of transformative technologies such as automobiles and aviation.

Adolescent giraffes, still mastering the dangers of acacias, humans must adapt to develop their defences and immune systems to extract necessary nourishment. Humans are refining strategies to interact with AI, evolving from initial efforts to sophisticated methods that manage Generative AI's ethical complexities. These strategies are expected to mature over time, mirroring the giraffe’s refined defenses, and enabling us to discern truth from AI-generated misinformation while adeptly navigating the landscape of AI ethics and governance. Addressing these issues requires an ongoing vigilant and proactive approach, comparable to the development of safety systems in response to the advent of transformative technologies such as automobiles and aviation. Understanding and regulating AI is a critical, continuous process that ensures its integration enhances societal well-being and upholds ethical standards.

However, given the emergent nature of Gen AI technology (i.e. capabilities of these systems continue to emerge and expand after deployment and during use),  many ethical implications like economic, economic and other societal impact/risk assessment and their mitigation remain open questions to resolve.

\subsection{Values Debt: Human Induced Ethical Flaws in Gen AI }

Human interactions with Generative AI, unlike natural systems, are driven by a broad spectrum of intentions, ranging from benign to malicious, and are shaped by a number of factors including the pursuit of innovation, power dynamics and market dominance. 
These complexities are pivotal in grasping the broader implications of AI on societal norms and individual behaviours, leading us into a discussion of \textbf{`Values Debt'} \cite{hussain2019values}—a concept that encapsulates the ethical and operational deficits incurred during the rapid advancement and deployment of Gen AI systems \cite{HouseOfCommons2024}. Similar to conventional software technical debt, both the principal and the ongoing interest must be paid until the debt is fully paid off \cite{ampatzoglou2020exploring}. 

\paragraph{Manifestation and Implications}
Generative AI systems often debut with significant biases, ethical issues, and inaccuracies, a stark contrast to the evolutionary precision observed in nature. These flaws frequently stem from the use of poor-quality or non-inclusive training datasets sourced from internet wikis, websites, blogs, and videos. The rush to capture market share, the pursuit of short-term financial gains, research agendas, and sometimes a lack of adequate measures or mere oversight contribute to these issues. Such a focus on immediate profits compromises the thoroughness of AI development and prioritizes short-term benefits over long-term safety and ethical considerations.

Consequently, this approach to AI development leads to the accumulation of `values debt,' where the pursuit of technological advancements incurs significant ethical and operational deficits that are difficult to rectify once the systems are deployed \cite{hussain2019values}. These deficits extend beyond mere technical issues, damaging the reputation, brand loyalty, and public image of developers. A notable example includes Google’s AI errors, such as recommending glue on pizza and asserting unfounded facts about celestial bodies—errors publicly acknowledged but described as unsolvable, at least for now, by the company’s leadership \cite{Walsh2024}. They encourage benefiting from the technology despite these known challenges, suggesting a similar 'tolerance or endurance' strategy discussed previously. 

In addressing the inherent challenges of misinformation and hallucinations in Generative AI, it is crucial to start with confronting manageable problems such as biased training data and premature model assumptions. This is more likely to align Gen AI outputs to human norms and societal expectations or values. While refining data input methods and enhancing interaction protocols can significantly impact AI accuracy and fairness, it is essential to recognize that these measures alone cannot solve all issues related to misinformation. However, these focused efforts are crucial for mitigating preventable risks and tackling more complex, intrinsic challenges.

\section{Recommendations: Fixing Gen AI's Value Alignment}

Generative AI profoundly influences societal norms and values, placing a significant ethical responsibility on its designers. As architects of future human conditions, designers are empowered to foster social change and bear immense moral responsibilities \cite{hussain2018integrating}. However, effectively aligning Gen AI with human values remains a critical challenge, evidenced by growing instances of value misalignment across various sectors (AI incident database, \cite{gabriel2024ethics, mcgregor2021preventing}).

One of the most promising frameworks proposed to address this challenge is the HHH framework by Askell et al. (2021)\cite{askell2021general}, which emphasizes that AI assistants should be helpful, honest, and harmless. These three principles serve as guiding pillars for AI development:
- \textbf{\textit{Helpfulness:}} AI should efficiently and concisely respond to all non-harmful queries, guiding users appropriately.
\textit{\textbf{- Honesty:}} AI must provide accurate information and be transparent about its capabilities and express appropriate levels of uncertainty without misleading human users. 
\textit{\textbf{- Harmlessness:}} AI should avoid aiding in dangerous acts (e.g. building a bomb) or causing offence or harm directly or indirectly through (any modality of) its output, and be considerate of cultural sensitivities and ethical implications in its interactions.

To operationalize these principles, Gabriel et al. 2024 \cite{gabriel2024ethics} propose a \textit{\textbf{tetradic relationship}} framework that involves viewing the relationship among the AI agent, user, developer, and society through a tetradic framework. This perspective advocates for a balanced AI system that satisfies the moral claims of all relevant parties without disproportionately favouring any single group. For instance, the system should not favour the AI agent or developer at the expense of the user or society, nor should it enable user actions that detrimentally affect societal welfare. However, as noted by the authors of HHH framework, this can not be achieved without some obvious trade-offs \cite{askell2021general} for example AI is conflicted ``between helpfulness to the user and harmlessness to others if agents are asked to aid in harmful activities." 

Nevertheless, the HHH framework can be quite effective when used in conjunction with the tetradic model, guiding the designers of Gen AI to help them adhere to ethical standards and promote a well-functioning sociotechnical system through responsible innovation. This approach challenges developers to rethink AI design, urging them to integrate these values from the onset to prevent several downstream \textit{\textbf{`Values Debt'}} \cite{hussain2019values} manifestations and ensure that AI developments are beneficial and sustainable for all stakeholders involved.

The adoption of the HHH framework, complemented by the tetradic relationship model, provides a robust methodology for aligning AI systems with societal values that remains a considerable challenge \cite{HouseOfCommons2024, touvron2023llama}. It requires a deep understanding of the interactions between AI technologies and societal structures and a commitment to ongoing evaluation and adaptation to ensure these systems remain aligned with evolving human values.

However, given the complexity of Gen AI, and its massive scale to engage and impact various stakeholders in our societies (including developers, researchers, policymakers, and users, mitigation efforts have to work in concert and continuously to overcome its challenges (at least most of them).

{\section{Conclusion}

This article has traversed the metaphorical Savannah, drawing parallels between the complex interactions of giraffes and acacias and the evolving relationship between humans and Generative AI. Through this analogy, we have explored the landscape of current risks and mitigation strategies akin to how adolescent giraffes navigate acacia defences. This formative phase of understanding and shaping generative AI addresses ethical and operational challenges through ongoing feedback, innovation, evaluation, governance, and regulation.

We highlight `Values Debt' of Gen AI as the additional cost for rapid and thoughtful AI development and a myriad of human motives, emphasizing the need for thoroughness to avert long-term ethical and operational pitfalls. The ‘HHH’ framework and the tetradic relationship model have identified pathways to reduce this debt by embedding values of helpfulness, honesty, and harmlessness in Gen AI agents, advancing the development of safety-aligned agents that resonate with human values.
This paper draws on the resilience found in both the natural world and human technological history to emphasize tolerance and endurance as critical strategies for navigating the challenges of Generative AI. Embracing temporary discomfort is key to unlocking significant long-term benefits, mirroring adaptive strategies that have enabled both natural systems and human societies to flourish amidst adversity.

With proactive governance and careful stewardship, we may be well-positioned to guide the development of Gen AI in ways that enhance societal well-being. This commitment ensures that our technological progress is matched by ethical advancements, fostering a symbiotic relationship where, akin to the mutual influence between giraffes and acacias, humans and AI continually shape each other for mutual benefit.


\bibliographystyle{ieeetr} 
\bibliography{bibliography.bib}

\begin{thebibliography}{100}

\bibitem{achiam2023gpt}
J.~Achiam, S.~Adler, S.~Agarwal, L.~Ahmad, I.~Akkaya, F.~L. Aleman, D.~Almeida, J.~Altenschmidt, S.~Altman, S.~Anadkat, {\em et~al.}, ``Gpt-4 technical report,'' {\em arXiv preprint arXiv:2303.08774}, 2023.

\bibitem{saetra2023generative}
H.~S. S{\ae}tra, ``Generative ai: Here to stay, but for good?,'' {\em Technology in Society}, vol.~75, p.~102372, 2023.

\bibitem{team2023gemini}
G.~Team, R.~Anil, S.~Borgeaud, Y.~Wu, J.-B. Alayrac, J.~Yu, R.~Soricut, J.~Schalkwyk, A.~M. Dai, A.~Hauth, {\em et~al.}, ``Gemini: a family of highly capable multimodal models,'' {\em arXiv preprint arXiv:2312.11805}, 2023.

\bibitem{chung2024scaling}
H.~W. Chung, L.~Hou, S.~Longpre, B.~Zoph, Y.~Tay, W.~Fedus, Y.~Li, X.~Wang, M.~Dehghani, S.~Brahma, {\em et~al.}, ``Scaling instruction-finetuned language models,'' {\em Journal of Machine Learning Research}, vol.~25, no.~70, pp.~1--53, 2024.

\bibitem{gabriel2024ethics}
I.~Gabriel, A.~Manzini, G.~Keeling, L.~A. Hendricks, V.~Rieser, H.~Iqbal, N.~Toma{\v{s}}ev, I.~Ktena, Z.~Kenton, M.~Rodriguez, {\em et~al.}, ``The ethics of advanced ai assistants,'' {\em arXiv preprint arXiv:2404.16244}, 2024.

\bibitem{ouyang2022training}
L.~Ouyang, J.~Wu, X.~Jiang, D.~Almeida, C.~Wainwright, P.~Mishkin, C.~Zhang, S.~Agarwal, K.~Slama, A.~Ray, {\em et~al.}, ``Training language models to follow instructions with human feedback,'' {\em Advances in neural information processing systems}, vol.~35, pp.~27730--27744, 2022.

\bibitem{Walsh2024}
T.~Walsh, ``Eat a rock a day, put glue on your pizza: How google’s ai is losing touch with reality,'' May 2024.
\newblock Accessed: 2024-06-05.

\bibitem{reid2024aioverviews}
L.~Reid, ``{AI Overviews: About last week}.'' \url{https://blog.google/products/search/ai-overviews-update-may-2024/}, 2024.
\newblock Accessed: 2025-06-05.

\bibitem{ferrara2024genai}
E.~Ferrara, ``Genai against humanity: Nefarious applications of generative artificial intelligence and large language models,'' {\em Journal of Computational Social Science}, pp.~1--21, 2024.

\bibitem{huang2024build}
K.~Huang, J.~Yeoh, S.~Wright, and H.~Wang, ``Build your security program for genai,'' in {\em Generative AI Security: Theories and Practices}, pp.~99--132, Springer, 2024.

\bibitem{hacker2023regulating}
P.~Hacker, A.~Engel, and M.~Mauer, ``Regulating chatgpt and other large generative ai models,'' in {\em Proceedings of the 2023 ACM Conference on Fairness, Accountability, and Transparency}, pp.~1112--1123, 2023.

\bibitem{laux2024trustworthy}
J.~Laux, S.~Wachter, and B.~Mittelstadt, ``Trustworthy artificial intelligence and the european union ai act: On the conflation of trustworthiness and acceptability of risk,'' {\em Regulation \& Governance}, vol.~18, no.~1, pp.~3--32, 2024.

\bibitem{dixon2023principled}
R.~B.~L. Dixon, ``A principled governance for emerging ai regimes: lessons from china, the european union, and the united states,'' {\em AI and Ethics}, vol.~3, no.~3, pp.~793--810, 2023.

\bibitem{mokander2023auditing}
J.~M{\"o}kander, J.~Schuett, H.~R. Kirk, and L.~Floridi, ``Auditing large language models: a three-layered approach,'' {\em AI and Ethics}, pp.~1--31, 2023.

\bibitem{cui2024risk}
T.~Cui, Y.~Wang, C.~Fu, Y.~Xiao, S.~Li, X.~Deng, Y.~Liu, Q.~Zhang, Z.~Qiu, P.~Li, {\em et~al.}, ``Risk taxonomy, mitigation, and assessment benchmarks of large language model systems,'' {\em arXiv preprint arXiv:2401.05778}, 2024.

\bibitem{ferrari2015writing}
R.~Ferrari, ``Writing narrative style literature reviews,'' {\em Medical writing}, vol.~24, no.~4, pp.~230--235, 2015.

\bibitem{rother2007systematic}
E.~T. Rother, ``Systematic literature review x narrative review,'' {\em Acta paulista de enfermagem}, vol.~20, pp.~v--vi, 2007.

\bibitem{grant2009typology}
M.~J. Grant and A.~Booth, ``A typology of reviews: an analysis of 14 review types and associated methodologies,'' {\em Health Information \& Libraries Journal}, vol.~26, no.~2, pp.~91--108, 2009.

\bibitem{dennett2013intuition}
D.~C. Dennett, {\em Intuition Pumps and Other Tools for Thinking}.
\newblock WW Norton \& Company, 2013.

\bibitem{dagg1971giraffa}
A.~I. Dagg, ``Giraffa camelopardalis,'' {\em Mammalian Species}, no.~5, pp.~1--8, 1971.

\bibitem{abdulrazak2000nutritive}
S.~Abdulrazak, T.~Fujihara, J.~Ondiek, and E.~{\O}rskov, ``Nutritive evaluation of some acacia tree leaves from kenya,'' {\em Animal Feed Science and Technology}, vol.~85, no.~1-2, pp.~89--98, 2000.

\bibitem{milewski1991thorns}
A.~V. Milewski, T.~P. Young, and D.~Madden, ``Thorns as induced defenses: experimental evidence,'' {\em Oecologia}, vol.~86, pp.~70--75, 1991.

\bibitem{thomas1993species}
R.~Thomas~Palo, J.~Gowda, and P.~H{\"o}gberg, ``Species height and root symbiosis, two factors influencing antiherbivore defense of woody plants in east african savanna,'' {\em Oecologia}, vol.~93, pp.~322--326, 1993.

\bibitem{madden1992symbiotic}
D.~Madden and T.~P. Young, ``Symbiotic ants as an alternative defense against giraffe herbivory in spinescent acacia drepanolobium,'' {\em Oecologia}, vol.~91, pp.~235--238, 1992.

\bibitem{zinn2007inducible}
A.~Zinn, D.~Ward, and K.~Kirkman, ``Inducible defences in acacia sieberiana in response to giraffe browsing,'' {\em African Journal of Range and Forage Science}, vol.~24, no.~3, pp.~123--129, 2007.

\bibitem{janzen1974swollen}
D.~H. Janzen, ``Swollen-thorn acacias of central america,'' {\em Smithsonian Contributions to Botany}, 1974.

\bibitem{perez2012anatomia}
W.~P{\'e}rez, V.~Michel, H.~Jerbi, and N.~Vazquez, ``Anatom{\'\i}a de la boca de la jirafa (giraffa camelopardalis rothschildi),'' {\em International Journal of Morphology}, vol.~30, no.~1, pp.~322--329, 2012.

\bibitem{agrawal1998dynamic}
A.~A. Agrawal and M.~T. Rutter, ``Dynamic anti-herbivore defense in ant-plants: the role of induced responses,'' {\em Oikos}, pp.~227--236, 1998.

\bibitem{mahenya2016giraffe}
O.~Mahenya, J.~K. Ndjamba, K.~M. Mathisen, and C.~Skarpe, ``Giraffe browsing in response to plant traits,'' {\em Acta Oecologica}, vol.~75, pp.~54--62, 2016.

\bibitem{wood2002scent}
W.~F. Wood and P.~J. Weldon, ``The scent of the reticulated giraffe (giraffa camelopardalis reticulata),'' {\em Biochemical systematics and ecology}, vol.~30, no.~10, pp.~913--917, 2002.

\bibitem{ogawa2018tannins}
S.~Ogawa and Y.~Yazaki, ``Tannins from acacia mearnsii de wild. bark: Tannin determination and biological activities,'' {\em Molecules}, vol.~23, no.~4, p.~837, 2018.

\bibitem{rehr1973chemical}
S.~Rehr, P.~P. Feeny, and D.~H. Janzen, ``Chemical defence in central american non-ant-acacias,'' {\em The Journal of Animal Ecology}, pp.~405--416, 1973.

\bibitem{furstenburg1994condensed}
D.~Furstenburg and W.~Van~Hoven, ``Condensed tannin as anti-defoliate agent against browsing by giraffe (giraffa camelopardalis) in the kruger national park,'' {\em Comparative Biochemistry and Physiology Part A: Physiology}, vol.~107, no.~2, pp.~425--431, 1994.

\bibitem{fernandez2008tongue}
L.~T. Fernandez, M.~J. Bashaw, R.~L. Sartor, N.~R. Bouwens, and T.~S. Maki, ``Tongue twisters: feeding enrichment to reduce oral stereotypy in giraffe,'' {\em Zoo Biology: Published in affiliation with the American Zoo and Aquarium Association}, vol.~27, no.~3, pp.~200--212, 2008.

\bibitem{clauss2003tannins}
M.~Clauss, A.~L. Fidgett, U.~Ganslosser, J.-M. Hatt, and J.~Nijboer, ``Tannins in the nutrition of wild animals: a review,'' in {\em Zoo Animal Nutrition Vol. II} (A.~L. Fidgett, M.~Clauss, U.~Ganslosser, J.-M. Hatt, and J.~Nijboer, eds.), pp.~53--89, Fürth: Filander Verlag, 2003.

\bibitem{pellew1984feeding}
R.~A. Pellew, ``The feeding ecology of a selective browser, the giraffe (giraffa camelopardalis tippelskirchi),'' {\em Journal of Zoology}, vol.~202, no.~1, pp.~57--81, 1984.

\bibitem{danell1994effects}
K.~Danell, R.~Bergstr{\"o}m, and L.~Edenius, ``Effects of large mammalian browsers on architecture, biomass, and nutrients of woody plants,'' {\em Journal of Mammalogy}, vol.~75, no.~4, pp.~833--844, 1994.

\bibitem{pellew1983impacts}
R.~Pellew, ``The impacts of elephant, giraffe and fire upon the acacia tortilis woodlands of the serengeti,'' {\em African Journal of Ecology}, vol.~21, no.~1, pp.~41--74, 1983.

\bibitem{lundberg1990low}
P.~Lundberg and M.~Astrom, ``Low nutritive quality as a defense against optimally foraging herbivores,'' {\em The American Naturalist}, vol.~135, no.~4, pp.~547--562, 1990.

\bibitem{augner1995low}
M.~Augner, ``Low nutritive quality as a plant defence: effects of herbivore-mediated interactions,'' {\em Evolutionary Ecology}, vol.~9, pp.~605--616, 1995.

\bibitem{dutoit1990giraffe}
J.~T. DUTOIT, ``Giraffe feeding on acacia flowers: predation or pollination?,'' {\em African Journal of Ecology}, vol.~28, no.~1, pp.~63--68, 1990.

\bibitem{weidinger2021ethical}
L.~Weidinger, J.~Mellor, M.~Rauh, C.~Griffin, J.~Uesato, P.-S. Huang, M.~Cheng, M.~Glaese, B.~Balle, A.~Kasirzadeh, {\em et~al.}, ``Ethical and social risks of harm from language models,'' {\em arXiv preprint arXiv:2112.04359}, 2021.

\bibitem{weidinger2022taxonomy}
L.~Weidinger, J.~Uesato, M.~Rauh, C.~Griffin, P.-S. Huang, J.~Mellor, A.~Glaese, M.~Cheng, B.~Balle, A.~Kasirzadeh, {\em et~al.}, ``Taxonomy of risks posed by language models,'' in {\em Proceedings of the 2022 ACM Conference on Fairness, Accountability, and Transparency}, pp.~214--229, 2022.

\bibitem{bender2021dangers}
E.~M. Bender, T.~Gebru, A.~McMillan-Major, and S.~Shmitchell, ``On the dangers of stochastic parrots: Can language models be too big?,'' in {\em Proceedings of the 2021 ACM conference on fairness, accountability, and transparency}, pp.~610--623, 2021.

\bibitem{kim2024m}
S.~S. Kim, Q.~V. Liao, M.~Vorvoreanu, S.~Ballard, and J.~W. Vaughan, ``I'm not sure, but...": Examining the impact of large language models' uncertainty expression on user reliance and trust,'' {\em arXiv preprint arXiv:2405.00623}, 2024.

\bibitem{gao2023retrieval}
Y.~Gao, Y.~Xiong, X.~Gao, K.~Jia, J.~Pan, Y.~Bi, Y.~Dai, J.~Sun, and H.~Wang, ``Retrieval-augmented generation for large language models: A survey,'' {\em arXiv preprint arXiv:2312.10997}, 2023.

\bibitem{hu2024rag}
Y.~Hu and Y.~Lu, ``Rag and rau: A survey on retrieval-augmented language model in natural language processing,'' {\em arXiv preprint arXiv:2404.19543}, 2024.

\bibitem{bai2022constitutional}
Y.~Bai, S.~Kadavath, S.~Kundu, A.~Askell, J.~Kernion, A.~Jones, A.~Chen, A.~Goldie, A.~Mirhoseini, C.~McKinnon, {\em et~al.}, ``Constitutional ai: Harmlessness from ai feedback,'' {\em arXiv preprint arXiv:2212.08073}, 2022.

\bibitem{glaese2022improving}
A.~Glaese, N.~McAleese, M.~Tr{\k{e}}bacz, J.~Aslanides, V.~Firoiu, T.~Ewalds, M.~Rauh, L.~Weidinger, M.~Chadwick, P.~Thacker, {\em et~al.}, ``Improving alignment of dialogue agents via targeted human judgements,'' {\em arXiv preprint arXiv:2209.14375}, 2022.

\bibitem{skalse2022defining}
J.~Skalse, N.~Howe, D.~Krasheninnikov, and D.~Krueger, ``Defining and characterizing reward gaming,'' {\em Advances in Neural Information Processing Systems}, vol.~35, pp.~9460--9471, 2022.

\bibitem{clemmer2024precisedebias}
C.~Clemmer, J.~Ding, and Y.~Feng, ``Precisedebias: An automatic prompt engineering approach for generative ai to mitigate image demographic biases,'' in {\em Proceedings of the IEEE/CVF Winter Conference on Applications of Computer Vision}, pp.~8596--8605, 2024.

\bibitem{gonzalez2024mitigating}
R.~Gonz{\'a}lez-Sendino, E.~Serrano, and J.~Bajo, ``Mitigating bias in artificial intelligence: Fair data generation via causal models for transparent and explainable decision-making,'' {\em Future Generation Computer Systems}, 2024.

\bibitem{bail2024can}
C.~A. Bail, ``Can generative ai improve social science?,'' {\em Proceedings of the National Academy of Sciences}, vol.~121, no.~21, p.~e2314021121, 2024.

\bibitem{10.5555/3618408.3619652}
S.~Santurkar, E.~Durmus, F.~Ladhak, C.~Lee, P.~Liang, and T.~Hashimoto, ``Whose opinions do language models reflect?,'' in {\em Proceedings of the 40th International Conference on Machine Learning}, ICML'23, JMLR.org, 2023.

\bibitem{mittelstadt2023unfairness}
B.~Mittelstadt, S.~Wachter, and C.~Russell, ``The unfairness of fair machine learning: Levelling down and strict egalitarianism by default,'' {\em arXiv preprint arXiv:2302.02404}, 2023.

\bibitem{wang2023survey}
Y.~Wang, Y.~Pan, M.~Yan, Z.~Su, and T.~H. Luan, ``A survey on chatgpt: Ai-generated contents, challenges, and solutions,'' {\em arXiv preprint arXiv:2305.18339}, 2023.

\bibitem{phuong2024evaluating}
M.~Phuong, M.~Aitchison, E.~Catt, S.~Cogan, A.~Kaskasoli, V.~Krakovna, D.~Lindner, M.~Rahtz, Y.~Assael, S.~Hodkinson, H.~Howard, T.~Lieberum, R.~Kumar, M.~Abi~Raad, A.~Webson, L.~Ho, S.~Lin, S.~Farquhar, M.~Hutter, G.~Del{\'e}tang, A.~Ruoss, S.~El-Sayed, S.~Brown, A.~Dragan, R.~Shah, A.~Dafoe, and T.~Shevlane, ``Evaluating frontier models for dangerous capabilities,'' {\em arXiv preprint arXiv:2403.13793}, 2024.

\bibitem{ausgov2023genai}
{Digital Transformation Agency} and {Department of Industry, Science and Resources}, ``Interim guidance on government use of public generative ai tools.'' Digital Transformation Agency, July 2023.
\newblock Accessed: 2024-05-29.

\bibitem{shevlane2023model}
T.~Shevlane, S.~Farquhar, B.~Garfinkel, M.~Phuong, J.~Whittlestone, J.~Leung, D.~Kokotajlo, N.~Marchal, M.~Anderljung, N.~Kolt, {\em et~al.}, ``Model evaluation for extreme risks,'' {\em arXiv preprint arXiv:2305.15324}, 2023.

\bibitem{golda2024privacy}
A.~Golda, K.~Mekonen, A.~Pandey, A.~Singh, V.~Hassija, V.~Chamola, and B.~Sikdar, ``Privacy and security concerns in generative ai: A comprehensive survey,'' {\em IEEE Access}, 2024.

\bibitem{greenfield2023social}
D.~Greenfield and S.~Bhavnani, ``Social media: generative ai could harm mental health,'' {\em Nature}, vol.~617, no.~7962, p.~676, 2023.

\bibitem{siegel2023weapons}
D.~Siegel and M.~B. Doty, ``Weapons of mass disruption: Artificial intelligence and the production of extremist propaganda,'' {\em Global Network on Extremism \& Technology, last modified February}, vol.~17, 2023.

\bibitem{lukas2023analyzing}
N.~Lukas, A.~Salem, R.~Sim, S.~Tople, L.~Wutschitz, and S.~Zanella-B{\'e}guelin, ``Analyzing leakage of personally identifiable information in language models,'' in {\em 2023 IEEE Symposium on Security and Privacy (SP)}, pp.~346--363, IEEE, 2023.

\bibitem{ozmen2023six}
O.~Ozmen~Garibay, B.~Winslow, S.~Andolina, M.~Antona, A.~Bodenschatz, C.~Coursaris, G.~Falco, S.~M. Fiore, I.~Garibay, K.~Grieman, {\em et~al.}, ``Six human-centered artificial intelligence grand challenges,'' {\em International Journal of Human--Computer Interaction}, vol.~39, no.~3, pp.~391--437, 2023.

\bibitem{brynjolfsson2023generative}
E.~Brynjolfsson, D.~Li, and L.~R. Raymond, ``Generative ai at work,'' tech. rep., National Bureau of Economic Research, 2023.

\bibitem{mannuru2023artificial}
N.~R. Mannuru, S.~Shahriar, Z.~A. Teel, T.~Wang, B.~D. Lund, S.~Tijani, C.~O. Pohboon, D.~Agbaji, J.~Alhassan, J.~Galley, {\em et~al.}, ``Artificial intelligence in developing countries: The impact of generative artificial intelligence (ai) technologies for development,'' {\em Information Development}, p.~02666669231200628, 2023.

\bibitem{wei2022emergent}
J.~Wei, Y.~Tay, R.~Bommasani, C.~Raffel, B.~Zoph, S.~Borgeaud, D.~Yogatama, M.~Bosma, D.~Zhou, D.~Metzler, {\em et~al.}, ``Emergent abilities of large language models,'' {\em arXiv preprint arXiv:2206.07682}, 2022.

\bibitem{schaeffer2024emergent}
R.~Schaeffer, B.~Miranda, and S.~Koyejo, ``Are emergent abilities of large language models a mirage?,'' {\em Advances in Neural Information Processing Systems}, vol.~36, 2024.

\bibitem{rillig2023risks}
M.~C. Rillig, M.~{\AA}gerstrand, M.~Bi, K.~A. Gould, and U.~Sauerland, ``Risks and benefits of large language models for the environment,'' {\em Environmental Science \& Technology}, vol.~57, no.~9, pp.~3464--3466, 2023.

\bibitem{simmons2024interoperable}
J.~C. Simmons and J.~M. Winograd, ``Interoperable provenance authentication of broadcast media using open standards-based metadata, watermarking and cryptography,'' {\em arXiv preprint arXiv:2405.12336}, 2024.

\bibitem{Wang_2023}
Y.~Wang, Y.~Pan, M.~Yan, Z.~Su, and T.~H. Luan, ``A survey on chatgpt: Ai–generated contents, challenges, and solutions,'' {\em IEEE Open Journal of the Computer Society}, vol.~4, p.~280–302, 2023.

\bibitem{contentcredentials2024}
{Coalition for Content Provenance and Authenticity}, ``Content credentials,'' 2024.
\newblock Accessed: 2024-05-30.

\bibitem{novelli2024generative}
C.~Novelli, F.~Casolari, P.~Hacker, G.~Spedicato, and L.~Floridi, ``Generative ai in eu law: Liability, privacy, intellectual property, and cybersecurity,'' {\em arXiv preprint arXiv:2401.07348}, 2024.

\bibitem{dalrymple2024towards}
D.~Dalrymple, J.~Skalse, Y.~Bengio, S.~Russell, M.~Tegmark, S.~Seshia, S.~Omohundro, C.~Szegedy, B.~Goldhaber, N.~Ammann, {\em et~al.}, ``Towards guaranteed safe ai: A framework for ensuring robust and reliable ai systems,'' {\em arXiv preprint arXiv:2405.06624}, 2024.

\bibitem{mokander2023operationalising}
J.~M{\"o}kander and L.~Floridi, ``Operationalising ai governance through ethics-based auditing: an industry case study,'' {\em AI and Ethics}, vol.~3, no.~2, pp.~451--468, 2023.

\bibitem{HouseOfCommons2024}
{Science, Innovation and Technology Committee}, ``Governance of artificial intelligence (ai),'' Committee Report Third Report of Session 2023--24, House of Commons, 2024.
\newblock Recommendations to government. The Government has two months to respond.

\bibitem{touvron2023llama}
H.~Touvron, L.~Martin, K.~Stone, P.~Albert, A.~Almahairi, Y.~Babaei, N.~Bashlykov, S.~Batra, P.~Bhargava, S.~Bhosale, D.~Bikel, L.~Blecher, C.~C. Ferrer, M.~Chen, G.~Cucurull, D.~Esiobu, J.~Fernandes, J.~Fu, W.~Fu, B.~Fuller, C.~Gao, V.~Goswami, N.~Goyal, A.~Hartshorn, S.~Hosseini, R.~Hou, H.~Inan, M.~Kardas, V.~Kerkez, M.~Khabsa, I.~Kloumann, A.~Korenev, P.~S. Koura, M.-A. Lachaux, T.~Lavril, J.~Lee, D.~Liskovich, Y.~Lu, Y.~Mao, X.~Martinet, T.~Mihaylov, P.~Mishra, I.~Molybog, Y.~Nie, A.~Poulton, J.~Reizenstein, R.~Rungta, K.~Saladi, A.~Schelten, R.~Silva, E.~M. Smith, R.~Subramanian, X.~E. Tan, B.~Tang, R.~Taylor, A.~Williams, J.~X. Kuan, P.~Xu, Z.~Yan, I.~Zarov, Y.~Zhang, A.~Fan, M.~Kambadur, S.~Narang, A.~Rodriguez, R.~Stojnic, S.~Edunov, and T.~Scialom, ``Llama 2: Open foundation and fine-tuned chat models,'' 2023.

\bibitem{dutoit1990regrowth}
J.~T. DuToit, J.~P. Bryant, and K.~Frisby, ``Regrowth and palatability of acacia shoots following pruning by african savanna browsers,'' {\em Ecology}, vol.~71, no.~1, pp.~149--154, 1990.

\bibitem{mcgregor2021preventing}
S.~McGregor, ``Preventing repeated real world ai failures by cataloging incidents: The ai incident database,'' in {\em Proceedings of the AAAI Conference on Artificial Intelligence}, vol.~35, pp.~15458--15463, 2021.

\bibitem{wei2022ai}
M.~Wei and Z.~Zhou, ``Ai ethics issues in real world: Evidence from ai incident database,'' {\em arXiv preprint arXiv:2206.07635}, 2022.

\bibitem{culkin1967schoolman}
J.~M. Culkin, ``A schoolman's guide to marshall mcluhan,'' 1967.

\bibitem{barbieri2024form}
L.~Barbieri and M.~Muzzupappa, ``Form innovation: investigating the use of generative design tools to encourage creativity in product design,'' {\em International Journal of Design Creativity and Innovation}, pp.~1--20, 2024.

\bibitem{epstein2023art}
Z.~Epstein, A.~Hertzmann, I.~of~Human~Creativity, M.~Akten, H.~Farid, J.~Fjeld, M.~R. Frank, M.~Groh, L.~Herman, N.~Leach, {\em et~al.}, ``Art and the science of generative ai,'' {\em Science}, vol.~380, no.~6650, pp.~1110--1111, 2023.

\bibitem{daar2023ludwig}
G.~F. Daar, I.~N. Sudipa, and T.~Gunas, ``Ludwig wittgenstein's concept of language game,'' {\em English Language Education Journal (ELEJ)}, vol.~2, no.~1, pp.~62--72, 2023.

\bibitem{levy2024anthropic}
S.~Levy, ``Ai is a black box. anthropic figured out a way to look inside,'' {\em Wired}, May 2024.
\newblock Accessed: 2023-05-23.

\bibitem{rudin2019stop}
C.~Rudin, ``Stop explaining black box machine learning models for high stakes decisions and use interpretable models instead,'' {\em Nature machine intelligence}, vol.~1, no.~5, pp.~206--215, 2019.

\bibitem{lazar2024frontier}
S.~Lazar, ``Frontier ai ethics: Anticipating and evaluating the societal impacts of generative agents,'' {\em arXiv preprint arXiv:2404.06750}, 2024.

\bibitem{cheong2024safeguarding}
I.~Cheong, A.~Caliskan, and T.~Kohno, ``Safeguarding human values: rethinking us law for generative ai’s societal impacts,'' {\em AI and Ethics}, pp.~1--27, 2024.

\bibitem{lim2023generative}
W.~M. Lim, A.~Gunasekara, J.~L. Pallant, J.~I. Pallant, and E.~Pechenkina, ``Generative ai and the future of education: Ragnar{\"o}k or reformation? a paradoxical perspective from management educators,'' {\em The international journal of management education}, vol.~21, no.~2, p.~100790, 2023.

\bibitem{ferrara2310genai}
E.~Ferrara, ``Genai against humanity: Nefarious applications of generative artificial intelligence and large language models,'' {\em arXiv preprint arXiv:2310.00737}, 2023.

\bibitem{zhou2024generative}
E.~Zhou and D.~Lee, ``Generative artificial intelligence, human creativity, and art,'' {\em PNAS nexus}, vol.~3, no.~3, p.~pgae052, 2024.

\bibitem{okatta2024navigating}
C.~G. Okatta, F.~A. Ajayi, and O.~Olawale, ``Navigating the future: integrating ai and machine learning in hr practices for a digital workforce,'' {\em Computer Science \& IT Research Journal}, vol.~5, no.~4, pp.~1008--1030, 2024.

\bibitem{walkowiak2023generative}
E.~Walkowiak and T.~MacDonald, ``Generative ai and the workforce: What are the risks?,'' {\em Available at SSRN}, 2023.

\bibitem{adam2021ai}
M.~Adam, M.~Wessel, and A.~Benlian, ``Ai-based chatbots in customer service and their effects on user compliance,'' {\em Electronic Markets}, vol.~31, no.~2, pp.~427--445, 2021.

\bibitem{kirwan2023future}
B.~Kirwan, ``The future impact of digital assistants on aviation safety culture,'' {\em Human Interaction and Emerging Technologies (IHIET-AI 2023): Artificial Intelligence and Future Applications. Ahram, T., and Taiar, R.(Eds)}, vol.~70, pp.~77--87, 2023.

\bibitem{sellen2023rise}
A.~Sellen and E.~Horvitz, ``The rise of the ai co-pilot: Lessons for design from aviation and beyond,'' {\em arXiv preprint arXiv:2311.14713}, 2023.

\bibitem{bogg2021can}
A.~Bogg, S.~Birrell, M.~A. Bromfield, and A.~M. Parkes, ``Can we talk? how a talking agent can improve human autonomy team performance,'' {\em Theoretical Issues in Ergonomics Science}, vol.~22, no.~4, pp.~488--509, 2021.

\bibitem{heinrich2023don}
P.~Heinrich, D.~Hollerer, C.~Karthaus, and M.~Gellrich, ``“don’t drop the plane to fly the mic!”: Designing for modern radiotelephony education in general aviation,'' in {\em 21. Fachtagung Bildungstechnologien der GI Fachgruppe Bildungstechnologien (DELFI), 10. Fachtagung Hochschuldidaktik Informatik (HDI), Aachen, Deutschland, 11.-14. September 2023}, pp.~187--192, Gesellschaft f{\"u}r Informatik, 2023.

\bibitem{gosper2021understanding}
S.~Gosper, J.~R. Trippas, H.~Richards, F.~Allison, C.~Sear, S.~Khorasani, and F.~Mattioli, ``Understanding the utility of digital flight assistants: A preliminary analysis,'' in {\em Proceedings of the 3rd Conference on Conversational User Interfaces}, pp.~1--5, 2021.

\bibitem{holford2022ethical}
W.~D. Holford, ``An ethical inquiry of the effect of cockpit automation on the responsibilities of airline pilots: Dissonance or meaningful control?,'' {\em Journal of Business Ethics}, vol.~176, no.~1, pp.~141--157, 2022.

\bibitem{church2023emerging}
K.~Church, A.~Schoene, J.~E. Ortega, R.~Chandrasekar, and V.~Kordoni, ``Emerging trends: Unfair, biased, addictive, dangerous, deadly, and insanely profitable,'' {\em Natural Language Engineering}, vol.~29, no.~2, pp.~483--508, 2023.

\bibitem{williamson2024era}
S.~M. Williamson and V.~Prybutok, ``The era of artificial intelligence deception: Unraveling the complexities of false realities and emerging threats of misinformation,'' {\em Information}, vol.~15, no.~6, p.~299, 2024.

\bibitem{dodgson1993learning}
M.~Dodgson, ``Learning, trust, and technological collaboration,'' {\em Human relations}, vol.~46, no.~1, pp.~77--95, 1993.

\bibitem{malone1994interdisciplinary}
T.~W. Malone and K.~Crowston, ``The interdisciplinary study of coordination,'' {\em ACM Computing Surveys (CSUR)}, vol.~26, no.~1, pp.~87--119, 1994.

\bibitem{bostrom2002existential}
N.~Bostrom, ``Existential risks: Analyzing human extinction scenarios and related hazards,'' {\em Journal of Evolution and technology}, vol.~9, 2002.

\bibitem{hussain2019values}
W.~Hussain, ``Values debt is eating software,'' {\em IEEE Software blog. Retrieved on}, vol.~12, 2019.

\bibitem{ampatzoglou2020exploring}
A.~Ampatzoglou, N.~Mittas, A.-A. Tsintzira, A.~Ampatzoglou, E.-M. Arvanitou, A.~Chatzigeorgiou, P.~Avgeriou, and L.~Angelis, ``Exploring the relation between technical debt principal and interest: An empirical approach,'' {\em Information and Software Technology}, vol.~128, p.~106391, 2020.

\bibitem{hussain2018integrating}
W.~Hussain, D.~Mougouei, and J.~Whittle, ``Integrating social values into software design patterns,'' in {\em Proceedings of the international workshop on software fairness}, pp.~8--14, 2018.

\bibitem{askell2021general}
A.~Askell, Y.~Bai, A.~Chen, D.~Drain, D.~Ganguli, T.~Henighan, A.~Jones, N.~Joseph, B.~Mann, N.~DasSarma, {\em et~al.}, ``A general language assistant as a laboratory for alignment,'' {\em arXiv preprint arXiv:2112.00861}, 2021.

\end{thebibliography}






\end{document}